\pdfoutput=1

\documentclass[11pt]{article}

\usepackage[final]{acl}

\usepackage{times}
\usepackage{latexsym}
\usepackage{todonotes}
\usepackage{makecell}  
\usepackage{xcolor}    

\usepackage[T1]{fontenc}

\usepackage[utf8]{inputenc}

\usepackage{microtype}
\usepackage{enumitem}

\usepackage{inconsolata}
\usepackage{subfig}

\usepackage{graphicx}
\usepackage{amsfonts}
\usepackage{arydshln}
\usepackage{amsmath}
\usepackage{multirow}
\usepackage{xcolor}
\usepackage{booktabs}

\title{On the \textit{Generalization vs Fidelity Paradox} in Knowledge Distillation}

\author{Suhas Kamasetty Ramesh$^*$ \and Ayan Sengupta$^*$ \and Tanmoy Chakraborty\\
Department of Electrical Engineering\\
Indian Institute of Technology Delhi, India\\
\texttt{suhaskr@gmail.com, ayan.sengupta@ee.iitd.ac.in, tanchak@iitd.ac.in}
}

\begin{document}
\maketitle

\def\thefootnote{*}\footnotetext{These authors contributed equally to this work.}\def\thefootnote{\arabic{footnote}}

\begin{abstract}
Knowledge distillation (KD) is a key technique for compressing large language models into smaller ones while preserving performance. Despite the recent traction of KD research, its effectiveness for smaller language models (LMs) and the mechanisms driving knowledge transfer remain underexplored. In this work, we present the first large-scale empirical and statistical analysis of KD across models ranging from 0.5B to 7B parameters on 19 complex reasoning and instruction following tasks in a zero-shot setting. Our findings reveal that KD can improve the average performance of smaller models by up to $10\%$, with a peak task specific gain of $22\%$, while providing only marginal benefits ($\sim 1.3\%$) for larger models. 
Surprisingly, teacher performance has a minimal impact on student outcomes, while teacher task expertise impacts KD effectiveness. A correlation study indicates that smaller LMs benefit more from KD, whereas larger LMs show diminished gains. Additionally, we uncover a misalignment between improvements in student performance and reasoning fidelity, suggesting that while KD enhances accuracy, it does not always maintain the structured decision-making processes of the teacher. Our ablation study further highlights the importance of teacher signals and logit smoothing in influencing students' performance after distillation. Overall, our study offers a comprehensive empirical and statistical assessment of KD, highlighting both its benefits and trade-offs when distilling knowledge from larger to smaller LMs.

\end{abstract}

\section{Introduction}
The rapid advancement of pre-trained language models (LMs) has led to the development of large-scale language models that achieve state-of-the-art performance across various NLP tasks \citep{dubey2024llama, qwen2.5, deepseekai2024deepseekv3technicalreport}. However, deploying these large models presents significant challenges due to their high computational and memory requirements \citep{zhu2023survey}. \text{Model compression} has emerged as a crucial technique to mitigate these challenges by reducing model size while preserving performance. Among various compression techniques, \text{knowledge distillation} (KD) has gained significant attention as it enables a smaller student model to learn from a larger teacher model, maintaining strong performance with reduced resource demands \citep{deng2020model}. KD plays a pivotal role in making large language models (LLMs) more accessible, facilitating their deployment in resource-constrained environments.

Several knowledge distillation approaches have been proposed to improve the training of student model and enhance generalization. \citet{hinton2015distilling} introduced the concept of soft target distillation, where the student learns from the softened output probabilities of the teacher model. Subsequent studies extended this idea to sequence-level KD for language models, such as SeqKD \citep{kim2016sequence}, which aligns the student model with the teacher's output distributions. More recent variants, including MiniLLM \citep{gu2024minillm}, leverage reverse Kullback-Leibler (KL) divergence to improve the learning of student model  for generative language models. 
Despite these advancements, a comprehensive understanding of KD techniques for LLMs remains unexplored. Current research primarily focuses on task-specific distillation setups, often overlooking broader implications across diverse tasks. Moreover, the distribution mismatch between training and inference remains a persistent challenge, particularly in autoregressive language models \citep{agarwal2024policy}, necessitating further exploration of large-scale KD strategies. 

\begin{figure*}[!htb]
    \centering
    \subfloat[Performance on mathematical reasoning tasks]{\includegraphics[width=0.95\linewidth]{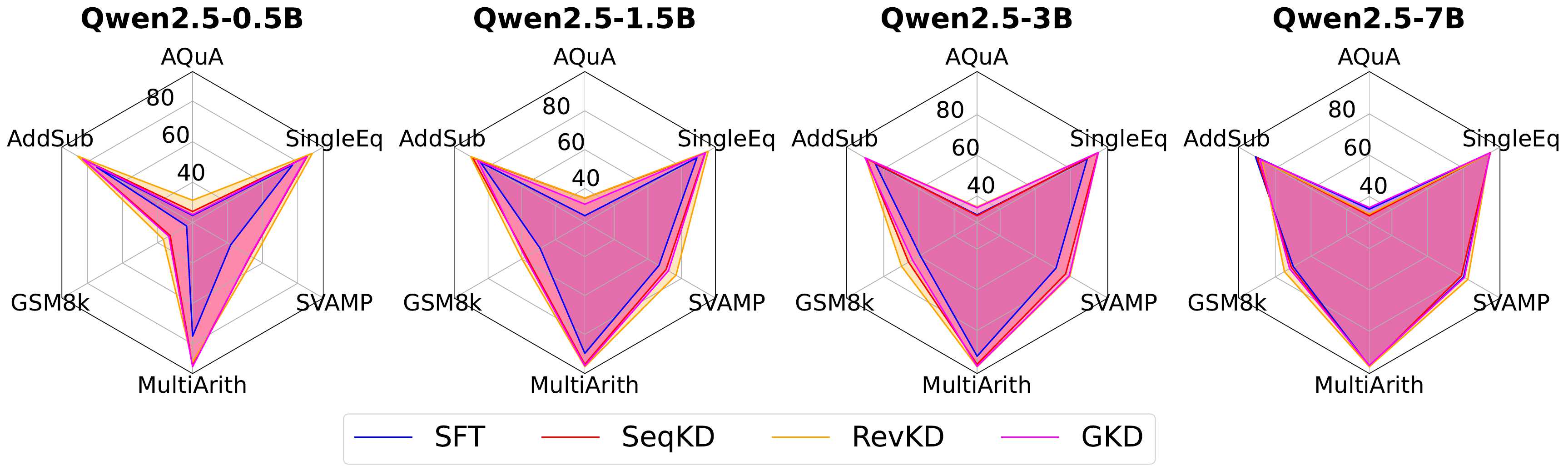}}
    \quad
    \subfloat[Performance on commonsense reasoning tasks]{    \includegraphics[width=0.95\linewidth]{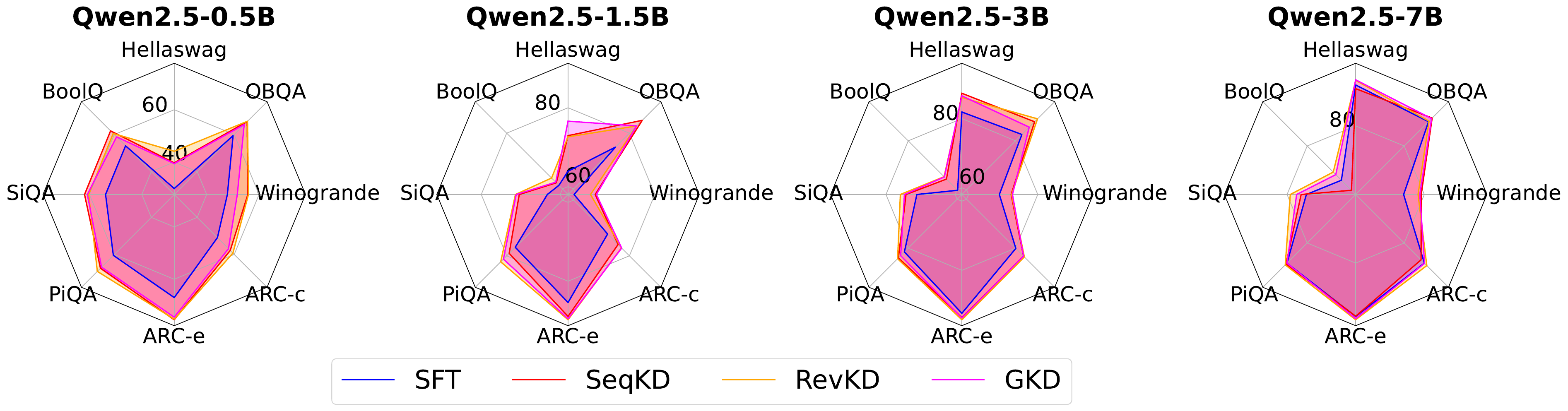}}
    \quad
    \subfloat[Performance on instruction following tasks]{    \includegraphics[width=0.93\linewidth]{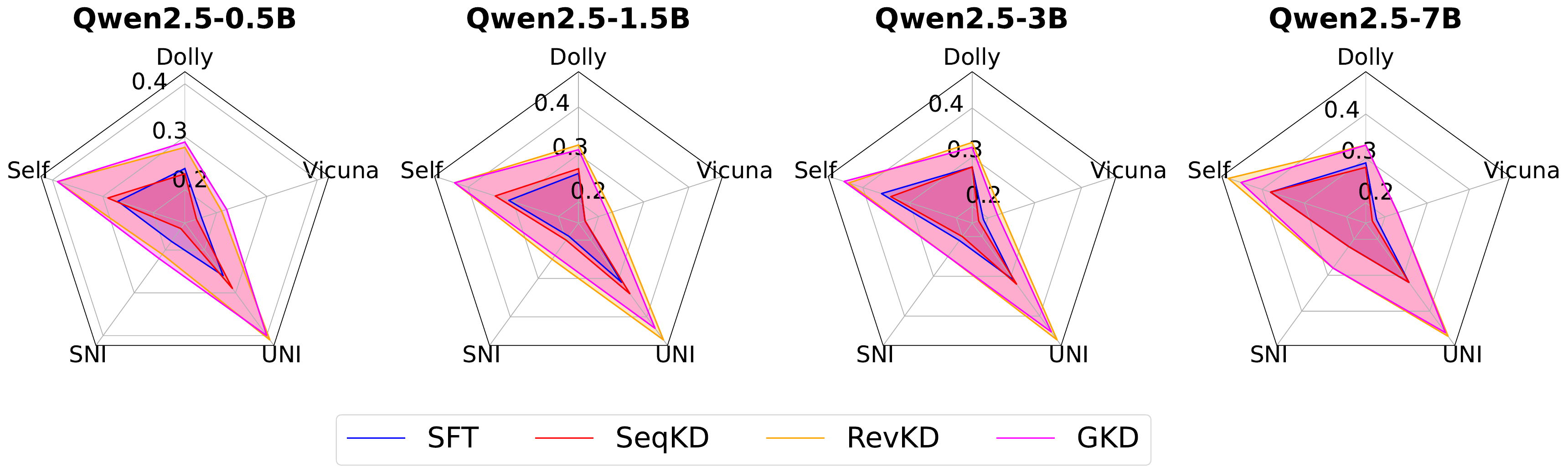}}
    \caption{Performance of different Qwen-2.5 student models across various (a) mathematical and (b) commonsense reasoning (c) instruction following tasks without and with distillation from larger Qwen models. We highlight the zero-shot accuracy for the supervised fine-tuned (SFT) model and the models distilled with different KD methods. The results are elaborated in Tables~\ref{tab:math_all_results}, ~\ref{tab:cs_all_results} and ~\ref{tab:dolly_all_results} in Appendix~\ref{appx:results}. Similar results for LLaMA student models are highlighted in Figure~\ref{fig:student_performance_llama} of Appendix~\ref{appx:results}.}
    \label{fig:student_performance}
\end{figure*}

In this paper, we conduct a large-scale empirical and statistical analysis to evaluate the effectiveness of knowledge distillation for small LMs ranging from 0.5B to 7B parameters across $19$ complex reasoning and instruction following tasks in a zero-shot setting. We investigate the impact of KD on the generalizability of small LMs to explore different aspects of knowledge distillation. We summarize the key findings from our study as follows\footnote{The source code of our study is made available at \url{https://github.com/LCS2-IIITD/KD_generalization}.}:

\begin{itemize}
[noitemsep,nolistsep,topsep=0pt,leftmargin=1em]
    \item \textbf{KD significantly improves model generalization.} Zero-shot performance of smaller LMs ($<1B$ size) can be improved by up to $10\%$, with peak gains reaching $22\%$ (c.f. Figure~\ref{fig:student_performance}) post-distillation from larger LMs ($7B-14B$).  
    \item \textbf{Larger models see diminishing returns.} 7B models exhibit only $\approx$1.3\% improvement after distillation, indicating that KD is most effective for smaller LMs.  
    \item \textbf{Choice of KD method has a marginal impact.} Despite varying improvement patterns, statistical analysis shows that different KD methods yield similar post-distillation performance. 
    \item \textbf{Teacher performance has minimal impact.} A Spearman rank test reveals that a stronger teacher does not necessarily lead to a better student. However, the teacher’s task expertise is crucial -- distilling from a task-unaware teacher can degrade student performance by up to 40\%.  
    \item \textbf{Smaller models benefit more from KD.} A high negative correlation exists between student model size and KD effectiveness, confirming that smaller LMs gain significantly more from distillation than larger ones.  
    \item \textbf{KD does not guarantee higher teacher-student agreement.} In mathematical reasoning tasks, the correlation between student performance and agreement with the teacher is statistically insignificant.  
    \item \textbf{High-performing students do not always exhibit reasoning fidelity.} Even when student models perform well, they do not necessarily replicate the teacher’s reasoning steps, highlighting a potential loss of structured decision-making.  
    \item \textbf{KD transfers task effectiveness but not always reasoning strategies.} The mismatch between student accuracy and poor fidelity suggests that while KD improves performance, it may fail to preserve the teacher’s structured decision-making process, raising concerns about interpretability and reliability in critical applications.  
\end{itemize}
%
These insights offer significant practical benefit to assess the benefits and trade-offs of knowledge distillation, encouraging researchers and practitioners in developing adaptive KD frameworks that balance knowledge transfer, reasoning integrity, and real-world applicability.

\section{Related Work}
\paragraph{KD for LLMs.} KD  has been widely used to compress LLMs while preserving performance \cite{hinton2015distilling}. Traditional KD relies on soft labels to transfer knowledge, with extensions such as feature-based \cite{romero2014fitnets}, self-distillation \cite{furlanello2018born}, multi-teacher distillation \cite{you2017learning} and collaborative distillation~\citep{sengupta2023good}. However, these approaches struggle with autoregressive sequence generation due to exposure bias and knowledge loss in smaller models. To address this, Generalized Knowledge Distillation trains students on self-generated sequences, mitigating distribution mismatch \cite{agarwal2024policy}. Instruction-tuning-CoT enables smaller models to inherit reasoning capabilities via instruction-based fine-tuning \cite{ranaldi2024aligning}. Multi-teacher KD aggregates strategies from multiple models, improving generalization \cite{tian2024beyond}, while MiniLLM prioritizes high-probability teacher outputs via reverse KL divergence, reducing overfitting \cite{gu2024minillm}. Adaptive Teaching KD (ATKD) adjusts knowledge transfer dynamically based on token difficulty, preventing degradation in large teacher-student setups \cite{zhong2024revisiting}. Despite advancements, most KD methods prioritize performance gains over understanding why certain strategies succeed or fail. Gaining deeper insights into KD effectiveness is crucial for developing more reliable, generalizable models.

\paragraph{Understanding effectiveness of KD.} While KD enhances model compression and transfer learning, its inner workings remain underexplored. Conventional KD methods often fail to preserve explainability, reducing trust in distilled models \cite{alharbi2021learning}. Studies challenge the assumption that high student-teacher fidelity ensures better generalization, revealing that students often struggle to mimic teachers due to optimization and dataset constraints \cite{stanton2021does}. KD can also transfer biases and adversarial vulnerabilities alongside useful knowledge \cite{ojha2023knowledge}. Attention and fidelity mechanisms are crucial, with research showing that diverse teacher attention maps enhance student generalization more effectively than rigid mimicry~\cite{guo2025does}. 
However, these studies primarily focus on vision tasks. Similar investigations are needed for language-based tasks to understand how KD influences reasoning, linguistic structures, and emergent capabilities of LLMs.

\paragraph{How our work is different?} Unlike prior works that primarily focus on task-specific knowledge distillation or performance benchmarking on a limited set of tasks, this study provides a large-scale empirical and statistical analysis of KD across diverse reasoning benchmarks with small LMs of varying sizes and capabilities. While most existing studies overlook explainability of KD, our work systematically examines the key factors influencing its effectiveness, providing valuable practical insights into the process.


\section{Methodology}
Here, we briefly describe three state-of-the-art knowledge distillation methods used in our study: Sequence-level knowledge distillation (SeqKD), Reverse KL knowledge distillation (RevKD), and Generalized Knowledge Distillation (GKD). Each method varies in its approach to training the student model and its associated loss objectives.

\paragraph{Sequence-level knowledge distillation.} SeqKD \citep{kim2016sequence} extends traditional KD~\citep{hinton2015distilling} by aligning the student’s output sequences with those of the teacher. Unlike standard KD, which operates at the token level, SeqKD forces the student model to generate entire sequences that match the teacher’s outputs. This is particularly useful for sequence generation tasks, such as machine translation and text summarization. The loss function for SeqKD is defined as:
\begin{equation}
\mathcal{L}_{\text{SeqKD}} = \mathbb{E}_{x \sim D} \left[ -\log P_S(y_T | x) \right],
\end{equation}
where $P_S(y_T | x)$ represents the probability of the teacher’s sequence output $y_T$ under the student model’s distribution. By training on full sequences, it improves the fluency and coherence of student-generated text compared to token-level distillation.

\paragraph{Reverse KL knowledge distillation.} \citet{gu2024minillm} proposed a novel KD strategy by minimizing the reverse KL divergence, instead of the standard forward KL divergence used in traditional KD. The motivation is to avoid overestimating low-probability regions of the teacher’s distribution, which is crucial for generative language models. The student model is trained using the following reverse KL loss:
\begin{equation}
\mathcal{L}_{\text{RevKD}} = \mathbb{E}_{x \sim D} \left[ D_{\text{KL}}(P_S(y|x) \parallel P_T(y|x)) \right],
\end{equation}
where $P_S(y|x)$ and $P_T(y|x)$ denote the logit distribution of student and teacher, respectively. 

\paragraph{Generalized knowledge distillation.} GKD~\citep{agarwal2024policy} addresses the distribution mismatch issue in traditional KD for auto-regressive sequence models. Standard KD methods train the student using a fixed dataset of teacher-generated or ground-truth sequences, leading to discrepancies between training and inference distributions. GKD mitigates this problem by incorporating an \textit{on-policy} learning approach, where the student model is trained on its self-generated sequences with feedback from the teacher. 
Given an input sequence \( x \), both the student \( P_S \) and teacher \( P_T \) generate output sequences. GKD mixes two data sources: (i) a fixed dataset containing ground-truth or teacher-generated sequences, and (ii) student-generated sequences, ensuring that the model learns from its own mistakes. The objective function of GKD is defined as:
\begin{align}
L_{\text{GKD}}(\theta) = & (1 - \lambda) \mathbb{E}_{(x,y) \sim (X,Y)} 
\left[ D(P_T || P_S(y | x) \right] \nonumber \\ 
& + \lambda \mathbb{E}_{x \sim X} \mathbb{E}_{y \sim p_S(\cdot | x)} 
\left[ D(P_T || P_S)(y | x) \right],
\end{align}
where \( D(P_T || P_S)(y | x) \) is a divergence measure between the teacher and student probability distributions, and \( \lambda \in [0,1] \) determines the fraction of on-policy student-generated data. GKD leverages the Jensen-Shannon Divergence (JSD), which interpolates between forward and reverse KL divergences using a mixing coefficient \( \beta \), given by:
\begin{align}
D_{\text{JSD}(\beta)}(P_T || P_S) = & \beta D_{\text{KL}}(P_T || M) \nonumber \\ 
& + (1 - \beta) D_{\text{KL}}(P_S || M),
\end{align}
where \( M = \beta P_T + (1 - \beta) P_S \) is the mixture distribution. When \( \beta \) approaches 0 or 1, JSD behaves like forward or reverse KL, respectively. By default, we use $\lambda = 0.5$ and $\beta=0.5$. This approach enables dynamic tuning of the divergence measure, optimizing the trade-off between generalization and generation diversity. 

Following~\citet{hinton2015distilling}, we use a temperature parameter $\tau$ to control the smoothness of the teacher and student token probabilities for all these KD methods. Under this assumption, the probability distribution of the teacher model is given by a temperature-scaled softmax function:
$P_T(y | x) = \frac{\exp(z_T / \tau)}{\sum_{i} \exp(z_i / \tau)}$, where $z_T$ represents the logits from the teacher model. The student model follows a similar formulation: $P_S(y | x) = \frac{\exp(z_S / \tau)}{\sum_{i} \exp(z_i / \tau)}$.


\newcommand{\mathhypone}
{
    \begin{tabular}{l|llllll|l}
    \cline{1-8}
    \textbf{Test Type} & \textbf{AQuA} & \textbf{AddSub} & \textbf{GSM8k} & \textbf{MultiArith} & \textbf{SVAMP} & \textbf{SingleEq} & \textbf{Average} \\
    \cline{1-8}
    t-test all methods & \textbf{4.07 (0.0)} & \textbf{5.57 (0.0)} & \textbf{8.87 (0.0)} & \textbf{7.09 (0.0)} & \textbf{7.39 (0.0)} & \textbf{8.31 (0.0)} & \textbf{8.47 (0.0)} \\
    \cdashline{1-8}
    t-test GKD & \textbf{2.16 (0.0)} & \textbf{2.80 (0.0)} & \textbf{5.72 (0.0)} & \textbf{3.81 (0.0)} & \textbf{4.88 (0.0)} & \textbf{5.37 (0.0)} & \textbf{5.10 (0.0)} \\
    t-test RevKD & \bf 4.05 (0.0) & \textbf{3.51 (0.0)} & \textbf{10.02 (0.0)} & \textbf{4.55 (0.0)} & \textbf{7.43 (0.0)} & \textbf{5.18 (0.0)} & \textbf{6.31 (0.0)} \\
    t-test SeqKD & 1.20 (0.1) & \textbf{3.36 (0.0)} & \textbf{3.04 (0.0)} & \textbf{4.12 (0.0)} & \textbf{2.30 (0.0)} & \textbf{3.84 (0.0)} & \textbf{3.72 (0.0)} \\
    \cdashline{1-8}
    ANOVA all methods & 2.67 (0.1) & 0.41 (0.7) & \textbf{5.62 (0.0)} & 0.52 (0.6) & \textbf{4.73 (0.0)} & 1.18 (0.3) & 2.27 (0.1) \\
    \cline{1-8}
    \end{tabular}%
}

\newcommand{\cshypone}
{
    \begin{tabular}{l|llllllll|l}
     \cline{1-10}
    \textbf{Test Type} & \textbf{ARC-c} & \textbf{ARC-e} & \textbf{BoolQ} & \textbf{Hellaswag} & \textbf{OBQA} & \textbf{PiQA} & \textbf{SiQA} & \textbf{Winogrande} & \textbf{Average} \\
     \cline{1-10}
    t-test all methods & \textbf{5.43 (0.0)} & \textbf{5.30 (0.0)} & 1.40 (0.1) & 0.66 (0.3) & \textbf{7.59 (0.0)} & \textbf{4.77 (0.0)} & \textbf{7.92 (0.0)} & \textbf{7.24 (0.0)} & \textbf{3.86 (0.0)} \\
    t-test GKD & \textbf{5.67 (0.0)} & \textbf{4.06 (0.0)} & \textbf{2.25 (0.0)} & 0.24 (0.4) & \textbf{4.90 (0.0)} & \textbf{3.49 (0.0)} & \textbf{6.64 (0.0)} & \textbf{4.16 (0.0)} & \textbf{2.96 (0.0)} \\
    t-test RevKD & \textbf{2.04 (0.0)} & \textbf{2.70 (0.0)} & 0.18 (0.4) & -0.04 (0.5) & \textbf{3.24 (0.0)} & \textbf{2.23 (0.0)} & \textbf{4.28 (0.0)} & \textbf{2.70 (0.0)} & 1.18 (0.1) \\
    t-test SeqKD & \textbf{3.25 (0.0)} & \textbf{2.48 (0.0)} & \textbf{1.96 (0.0)} & \textbf{3.13 (0.0)} & \textbf{5.74 (0.0)} & \textbf{2.90 (0.0)} & \textbf{3.37 (0.0)} & \textbf{11.34 (0.0)} & \textbf{4.33 (0.0)} \\
    \cdashline{1-10}
    ANOVA all methods & 0.49 (0.6) & 0.72 (0.5) & 0.20 (0.8) & 0.18 (0.8) & 0.04 (1.0) & 0.45 (0.6) & 1.12 (0.3) & 0.27 (0.8) & 0.30 (0.8) \\
     \cline{1-10}
    \end{tabular}%
}

\newcommand{\inthypone}
{
    \begin{tabular}{l|lllll|l}
    \cline{1-7}
    Test Type & Dolly & Self  & SNI   & UNI   & Vicuna & Average \\
    \cline{1-7}
    t-test all methods & \textbf{6.31 (0.0)} & \textbf{4.85 (0.0)} & \textbf{7.01 (0.0)} & \textbf{8.23 (0.0)} & \textbf{5.35 (0.0)} & \textbf{7.18 (0.0)} \\
    t-test GKD & \textbf{14.09 (0.0)} & \textbf{10.73 (0.0)} & \textbf{11.25 (0.0)} & \textbf{18.2 (0.0)} & \textbf{11.74 (0.0)} & \textbf{25.66 (0.0)} \\
    t-test RevKD & \textbf{16.63 (0.0)} & \textbf{8.21 (0.0)} & \textbf{15.69 (0.0)} & \textbf{35.42 (0.0)} & \textbf{12.17 (0.0)} & \textbf{34.02 (0.0)} \\
    t-test SeqKD & -2.63 (1.0) & -1.85 (0.9) & -0.32 (0.6) & 1.55 (0.1) & -3.63 (1.0) & -1.66 (0.9) \\
    \cdashline{1-7}
    ANOVA all methods & \textbf{117.82 (0.0)} & \textbf{40.05 (0.0)} & \textbf{67.28 (0.0)} & \textbf{196.88 (0.0)} & \textbf{85.16 (0.0)} & \textbf{329.47 (0.0)} \\
    \cline{1-7}
    \end{tabular}%
}

\begin{table*}[!htb]
  \centering
    \subfloat[Mathematical reasoning]{\scalebox{0.84}\mathhypone
\label{tab:hypo1math}}
\quad
    \subfloat[Commonsense reasoning]{\scalebox{0.7}\cshypone
\label{tab:hypo1cs}}
\quad
    \subfloat[Instruction following]{\scalebox{0.7}\inthypone
\label{tab:hypo1int}}
    \caption{Statistical t-test for understanding the statistical significance of KD on student models' performance. We calculate the t-statistics and $p$-value for all KD methods and for individually for each method. We further measure the ANOVA F-statistics to underscore the differences between different KD methods on the student performance. \textbf{Bold} indicates that the results are statistically significant ($p$-value $<0.05$).}
  \label{tab:hypo1}%
  \vspace{-3mm}
\end{table*}%

\section{Experimental Setup}
In our empirical study, we use Qwen-2.5~\cite{qwen2.5} (0.5B, 1.5B, 3B, 7B, 14B) and LLaMA-3~\cite{dubey2024llama} (3.2-1B, 3.2-3B, 3.1-8B) model series, with all pretrained weights sourced from Huggingface~\cite{wolf-etal-2020-transformers}, evaluated on mathematical reasoning, commonsense reasoning and instruction following tasks. For mathematical reasoning, we fine-tune base models on Math10K~\cite{hu2023llm} and evaluate on GSM8K~\cite{cobbe2021training}, SVAMP~\cite{patel2021nlp}, MultiArith~\cite{roy2015solving}, AddSub~\cite{hosseini2014learning}, AQuA~\cite{ling2017program}, and SingleEq~\cite{koncel2015parsing}. For commonsense reasoning, we fine-tune on Commonsense15K~\cite{hu2023llm} and evaluate on Hellaswag~\cite{zellers2019hellaswag}, Winogrande~\cite{sakaguchi2021winogrande}, ARC~\cite{allenai:arc}, OBQA~\cite{OpenBookQA2018}, BoolQ~\cite{clark2019boolq}, PiQA~\cite{Bisk2020}, and SiQA~\cite{sap-etal-2019-social}. 
For instruction following, we fine-tune using Dolly-15K~\cite{gu2024minillm} and evaluate on Dolly, SelfInst~\cite{selfinstruct} (denoted by `self' throughout the paper), Vicuna~\cite{vicuna2023}, SNI~\cite{wang-etal-2022-super} and UNI~\cite{honovich-etal-2023-unnatural}.
Detailed dataset descriptions and splits are provided in Appendix~\ref{appx:datasets} and Table~\ref{tab:data_split_float}. We use LoRA~\cite{hu2021lora} adapters with $r=8$ and $\alpha=16$ for supervised fine-tuning and KD fine-tuning of all the models. 
All the experiments were conducted on a single Nvidia-A100 GPU.

We use a batch-size of 16, learning rate of $3\times10^{-4}$ and max-length of 256 across all models and methods. We fine-tune the models for 3 and 4 epochs for Commonsense15K and Math10K datasets, respectively. Following~\citet{gu2024minillm}, for zero-shot evaluation we set $\tau = 1$ (default, for temperature ablation experiments highlighted in Figure~\ref{fig:temp_ablation} we use $\tau \in \{1, 2, 5 \}$), top-$p=1$, top-$k=0$ and $\text{num}\_\text{beams}=1$.

\newcommand{\mathhyptwo}
{
    \begin{tabular}{l|llllll|l}
     \cline{1-8}
    \textbf{Test Type} & \textbf{AQuA} & \textbf{AddSub} & \textbf{GSM8k} & \textbf{MultiArith} & \textbf{SVAMP} & \textbf{SingleEq} & \textbf{Average} \\
     \cline{1-8}
    Spearman rank & -0.10 (0.6) & 0.30 (0.1) & 0.09 (0.6) & 0.21 (0.2) & -0.09 (0.6) & \textbf{0.35 (0.0)} & -0.01 (1.0) \\
     \cline{1-8}
    \end{tabular}%
}

\newcommand{\cshyptwo}
{
    \begin{tabular}{l|llllllll|l}
     \cline{1-10}
    \textbf{Test Type} & \textbf{ARC-c} & \textbf{ARC-e} & \textbf{BoolQ} & \textbf{Hellaswag} & \textbf{OBQA} & \textbf{PiQA} & \textbf{SiQA} & \textbf{Winogrande} & \textbf{Average} \\
     \cline{1-10}
    Spearman rank & \textbf{-0.41 (0.0)} & \textbf{-0.36 (0.0)} & 0.1 (0.6) & -0.32 (0.1) & \textbf{-0.41 (0.0)} & -0.27 (0.1) & -0.31 (0.1) & -0.02 (0.9) & -0.3 (0.1)  \\
     \cline{1-10}
    \end{tabular}%
}

\newcommand{\inthyptwo}
{
    \begin{tabular}{l|lllll|l}
    \cline{1-7}
    \textbf{Test Type} & \textbf{Dolly} & \textbf{Self} & \textbf{SNI} & \textbf{UNI} & \textbf{Vicuna} & \textbf{Average} \\
    \cline{1-7}
    Spearman rank & 0.16 (0.4) & 0.04 (0.8) & 0.16 (0.4) & 0.03 (0.9) & -0.0 (1.0) & 0.14 (0.4) \\
    \cline{1-7}
    \end{tabular}%
}

\begin{table*}[!htb]
  \centering
    \subfloat[Mathematical reasoning]{\scalebox{0.8}\mathhyptwo
\label{tab:hypo2math}}
\quad
    \subfloat[Commonsense reasoning]{\scalebox{0.7}\cshyptwo
\label{tab:hypo2cs}}
\quad
    \subfloat[Instruction following]{\scalebox{0.7}\inthyptwo
\label{tab:hypo2int}}
        \caption{Spearman rank correlation and $p$-value between student and teacher performance.} 
  \label{tab:hypo2}%
\end{table*}%

\paragraph{Measures for quantifying teacher-student agreement and fidelity.} In this paper, along with the performance measurement of student models, we use \textit{teacher-student agreement} and \textit{reasoning fidelity} scores to understand alignment between teacher and student post-KD. Agreement quantifies how often student replicates the teacher’s outputs and is measured using top-1 agreement (fraction of matching predictions). However, this metric only relies on the final answer and does not measure the quality of the intermediate reasoning steps. Reasoning fidelity, on the other hand, captures how well the student mirrors the teacher’s reasoning process rather than just final predictions. We use BLEU~\cite{papineni2002bleu} score between teacher and student reasoning outputs to compute fidelity. 

\section{Experimental Results}
In this section, we elaborate the empirical and statistical results obtained in our study.

\begin{figure*}[htbp]
    \centering
    \subfloat[Mathematical reasoning]{\includegraphics[width=0.42\linewidth]{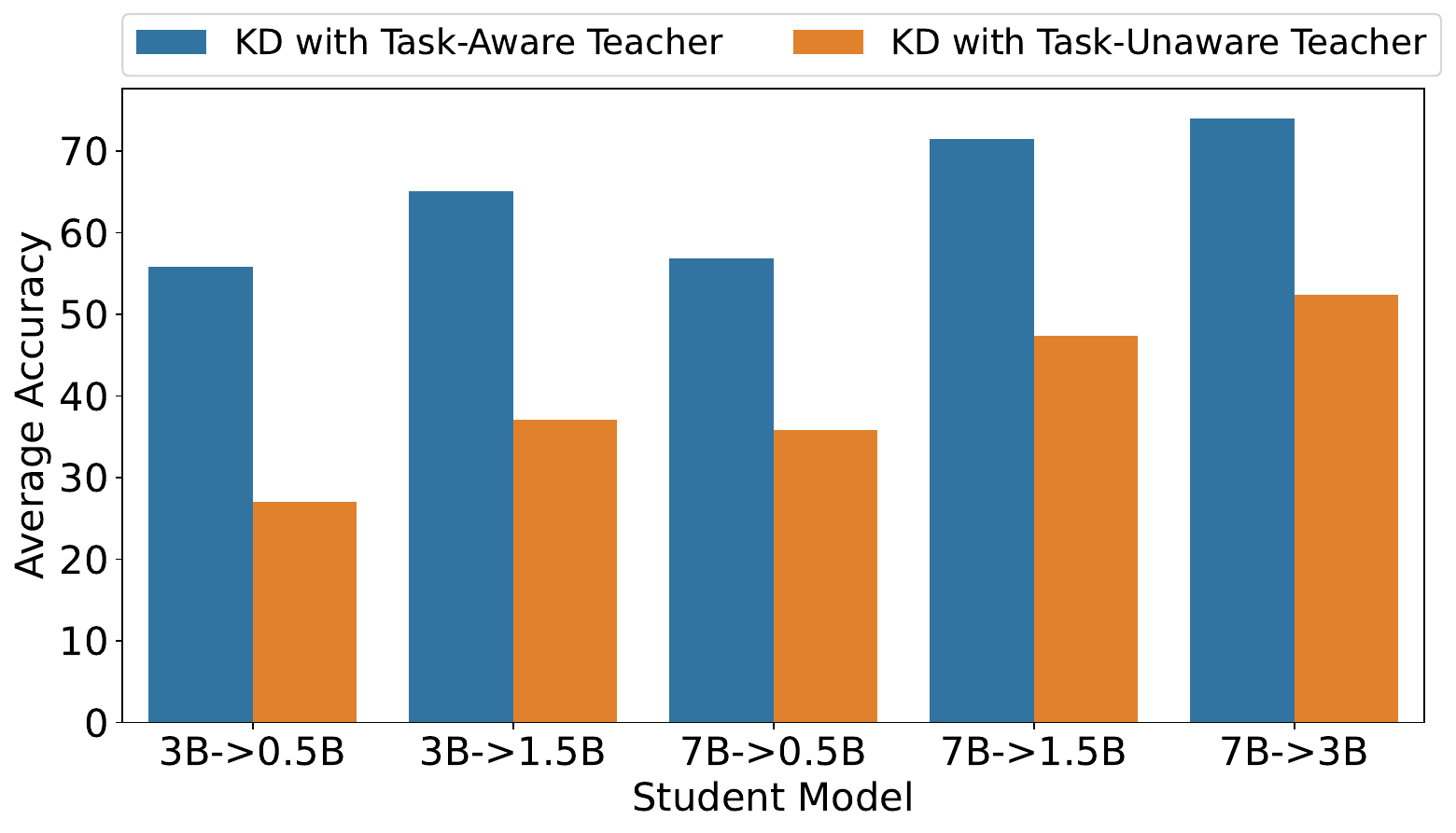}}
    \subfloat[Commonsense reasoning]{\includegraphics[width=0.42\linewidth]{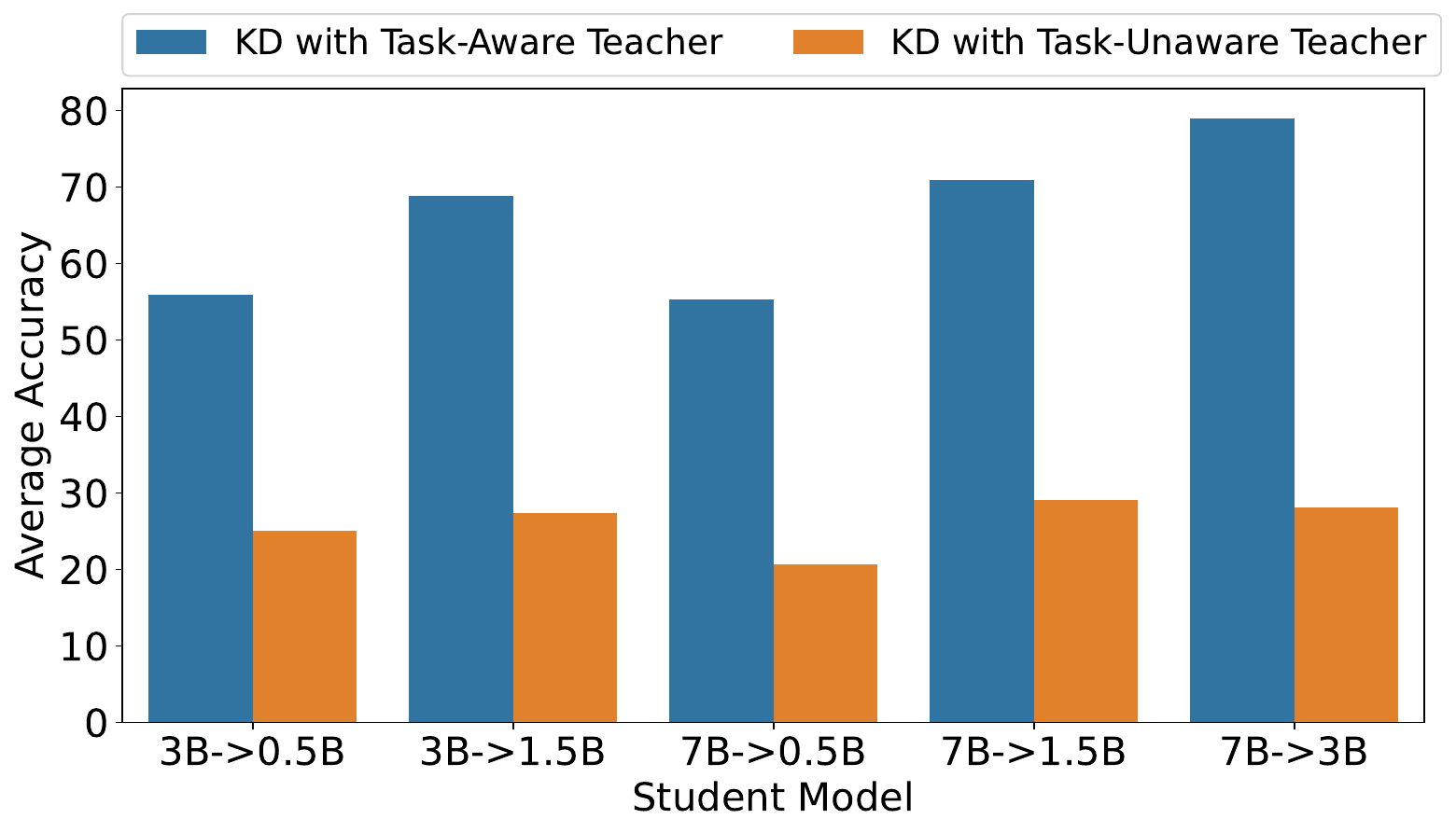}}
    \caption{Impact of teacher task adaptation on distillation effectiveness. We evaluate the student performance post-distillation with teacher without/with fine-tuning on mathematical and commonsense reasoning tasks.}
    \label{fig:teacher_ablation}
\end{figure*}

\newcommand{\mathhypthree}
{
    \begin{tabular}{l|llllll|l}
     \cline{1-8}
    \textbf{Test Type} & \textbf{AQuA} & \textbf{AddSub} & \textbf{GSM8k} & \textbf{MultiArith} & \textbf{SVAMP} & \textbf{SingleEq} & \textbf{Average} \\
     \cline{1-8}
    Spearman rank & -0.23 (0.2) & \textbf{-0.64 (0.0)} & \textbf{-0.51 (0.0)} & \textbf{-0.83 (0.0)} & \textbf{-0.43 (0.0)} & \textbf{-0.59 (0.0)} & \textbf{-0.66 (0.0)}\\
     \cline{1-8}
    \end{tabular}%
}

\newcommand{\cshypthree}
{
    \begin{tabular}{l|llllllll|l}
     \cline{1-10}
    \textbf{Test Type} & \textbf{ARC-c} & \textbf{ARC-e} & \textbf{BoolQ} & \textbf{Hellaswag} & \textbf{OBQA} & \textbf{PiQA} & \textbf{SiQA} & \textbf{Winogrande} & \textbf{Average} \\
     \cline{1-10}
    Spearman rank & \textbf{-0.63 (0.0)} & \textbf{-0.78 (0.0)} & -0.15 (0.4) & \textbf{-0.4 (0.0)} & \textbf{-0.55 (0.0)} & \textbf{-0.67 (0.0)} & \textbf{-0.49 (0.0)} & \textbf{-0.34 (0.0)} & \textbf{-0.54 (0.0)} \\
     \cline{1-10}
    \end{tabular}%
}

\newcommand{\inthypthree}
{
    \begin{tabular}{l|lllll|l}
    \cline{1-7}
    \textbf{Test Type} & \textbf{Dolly} & \textbf{Self} & \textbf{SNI} & \textbf{UNI} & \textbf{Vicuna} & \textbf{Average} \\
    \cline{1-7}
    Spearman rank & 0.08 (0.7) & 0.3 (0.1) & -0.32 (0.1) & -0.02 (0.9) & 0.12 (0.5) & -0.03 (0.9) \\
    \cline{1-7}
    \end{tabular}%
}

\begin{table*}[!htb]
  \centering
    \subfloat[Mathematical reasoning]{\scalebox{0.8}\mathhypthree
\label{tab:hypo3math}}
\quad
    \subfloat[Commonsense reasoning]{\scalebox{0.7}\cshypthree
\label{tab:hypo3cs}}
\quad
    \subfloat[Instruction following]{\scalebox{0.7}\inthypthree
\label{tab:hypo3int}}
        \caption{Spearman rank correlation and $p$-value between student performance and student model size.} 
  \label{tab:hypo3}%
  \vspace{-3mm}
\end{table*}%
\newcommand{\mathhypfour}
{
    \begin{tabular}{l|llllll}
     \cline{1-7}
    \textbf{Test Type} & \textbf{AQuA} & \textbf{AddSub} & \textbf{GSM8k} & \textbf{MultiArith} & \textbf{SVAMP} & \textbf{SingleEq} \\
     \cline{1-7}
    Spearman rank & -0.18 (0.3) & -0.09 (0.6) & -0.33 (0.1) & \textbf{-0.63 (0.0)} & \textbf{-0.37 (0.0)} & -0.12 (0.5) \\
     \cline{1-7}
    \end{tabular}%
}

\newcommand{\cshypfour}
{
    \begin{tabular}{l|llllllll}
     \cline{1-9}
    \textbf{Test Type} & \textbf{ARC-c} & \textbf{ARC-e} & \textbf{BoolQ} & \textbf{Hellaswag} & \textbf{OBQA} & \textbf{PiQA} & \textbf{SiQA} & \textbf{Winogrande} \\
     \cline{1-9}
    Spearman rank & -0.12 (0.5) & -0.19 (0.3) & 0.21 (0.2) & 0.08 (0.7) & -0.32 (0.1) & -0.02 (0.9) & -0.21 (0.2) & 0.12 (0.5) \\
     \cline{1-9}
    \end{tabular}%
}

\begin{table*}[!htb]
  \centering
    \subfloat[Mathematical reasoning]{\scalebox{0.8}\mathhypfour
\label{tab:hypo4math}}
\quad
    \subfloat[Commonsense reasoning]{\scalebox{0.75}\cshypfour
\label{tab:hypo4cs}}
        \caption{Spearman rank correlation and $p$-value between student performance and teacher-student agreement.} 
  \label{tab:hypo4}%
  \vspace{-3mm}
\end{table*}%

\paragraph{Effectiveness of KD.}  
Figure~\ref{fig:student_performance} shows that KD consistently outperforms SFT in all three benchmarks - mathematical reasoning, commonsense reasoning and instruction following , particularly benefiting smaller models. For Qwen-0.5B, KD improves performance by $10.4\%$ and $7.8\%$ for math and commonsense reasoning respectively and $0.08$ absolute points for instruction following, but this effect diminishes as student size increases ($7.5\%$, $5.8\%$, $0.09$ for Qwen-1.5B, $5.4\%$, $3.5\%$, $0.07$  for Qwen-3B and $0.2\%$, $1.9\%$, $0.08$ for Qwen-7B). Similar performance improvements are observed with LLaMA student models -- $12.0\%$, $12.7\%$ and $0.06$ for LLaMA-1B and $3.7\%$, $4.9\%$ and $0.07$ for LLaMA-3B model. Among KD methods, RevKD is the most consistent, yielding an average gain of $6.3\%$ across all student sizes. The statistical t-test results (Table~\ref{tab:hypo1}) confirm these improvements as significant ($p$-value $< 0.05$). While one-way ANOVA test reveals a difference between KD methods on instruction-following tasks, it shows no significant differences on mathematical and commonsense reasoning tasks—suggesting that all KD methods perform similarly on these two benchmarks.

\paragraph{Does KD depend on teacher performance?}  
Table~\ref{tab:hypo2} reports the Spearman rank correlation between student improvement after KD and teacher performance. In mathematical reasoning, structured tasks like AddSub (0.30, $p$-value=$0.1$) and SingleEq (0.35, $p$-value=$0.0$) show positive correlations, suggesting stronger teachers enhance student performance. However, complex reasoning tasks such as GSM8k ($0.09$, $p$-value=$0.6$) and AQuA ($-0.10$, $p$-value=$0.6$) exhibit weak or negative correlations, indicating that teacher quality alone does not dictate KD success. In commonsense reasoning and instruction following tasks, correlations are generally weak or negative, implying minimal influence of teacher performance. Overall, teacher quality inconsistently impacts student gains, with structured mathematical tasks benefiting more than open-ended reasoning tasks. However, Figure~\ref{fig:teacher_ablation} shows that a task-unaware teacher can significantly degrade student performance. Post-distillation performance of student model can drop up to $40\%$, if teacher is not fine-tuned on downstream domain.

\paragraph{Does KD depend on student model size?}  \label{analysis_kd_student_size}
Table~\ref{tab:hypo3} shows a strong negative correlation ($-0.66$, $p$-value=$0.0$) between KD improvement and student size, indicating diminishing returns as model size increases. This trend is most pronounced in MultiArith ($-0.83$, $p$-value=$0.0$), AddSub ($-0.64$, $p$-value=$0.0$) and SingleEq ($-0.59$, $p$-value=$0.0$), where larger models already possess strong reasoning capabilities. In commonsense reasoning, similar effect is observed ($-0.54$, $p$-value=$0.0$). In instruction following, correlations are mostly weak. These results confirm that KD benefits smaller LMs significantly, while larger LMs see diminishing returns as they already exhibit strong reasoning. 

\begin{figure*}[!htb]
    \centering
    \subfloat[Mathematical reasoning tasks]{
    \includegraphics[width=0.88\linewidth]{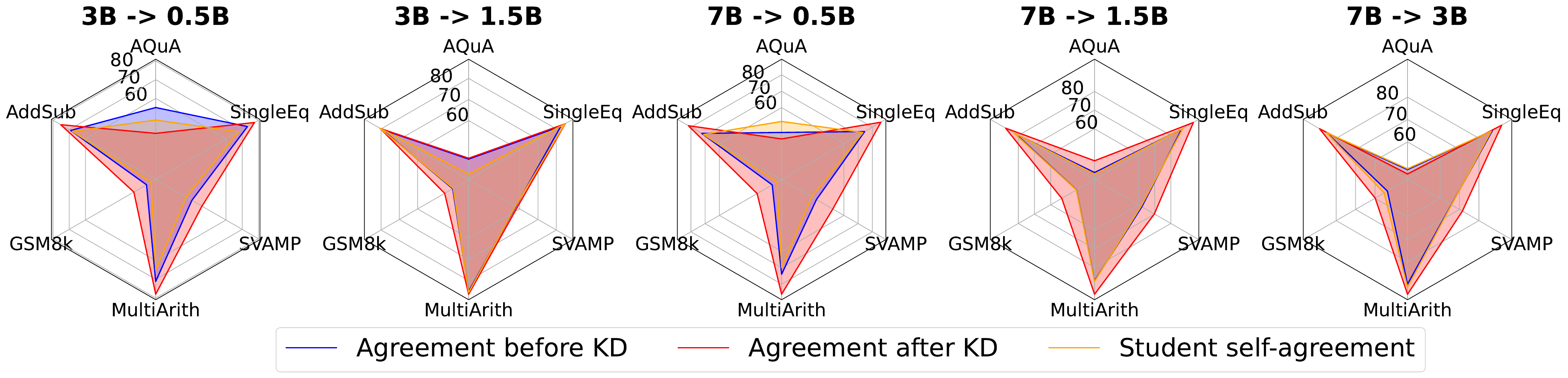}}
    \quad
    \subfloat[Commonsense reasoning tasks]{
    \includegraphics[width=0.88\linewidth]{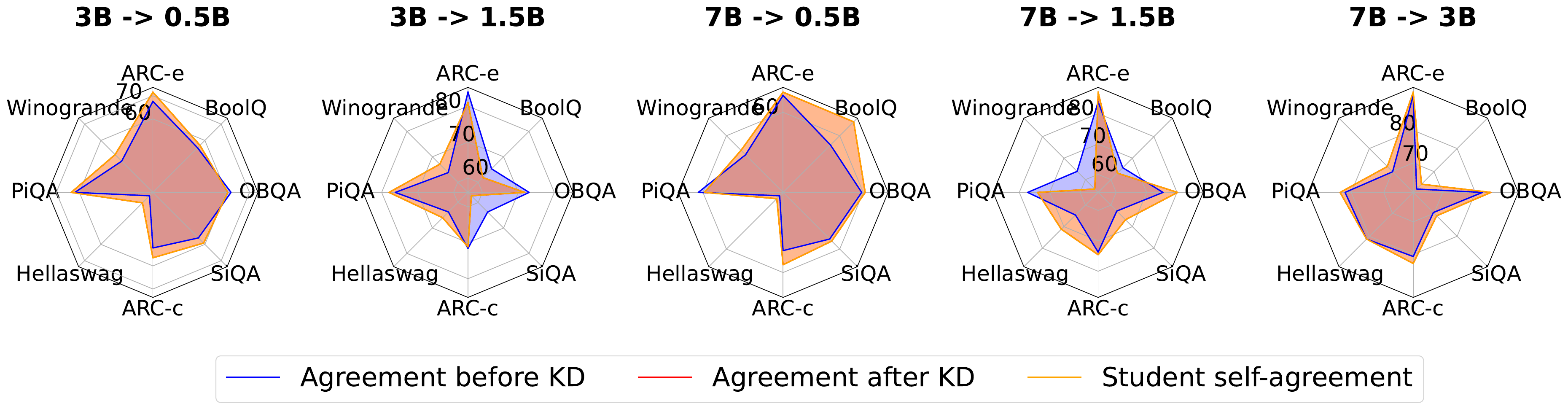}}
    \caption{Student agreement with teacher before KD and after KD on (a) mathematical and (b) commonsense reasoning tasks. We omit the instruction following tasks as the final output is in free-text form, therefore we can not determine exact match. We also highlight the student's self-agreement between the before and after KD outputs. Results with Qwen-14B and LLaMA-8B teachers are highlighted in Figures~\ref{fig:student_agreement2} and~\ref{fig:student_agreement_llama} of Appendix~\ref{appx:results}, respectively.}
    \label{fig:student_agreement}
\end{figure*}

\begin{figure*}[!htb]
    \centering
    \subfloat[Mathematical reasoning tasks]{\includegraphics[width=0.88\linewidth]{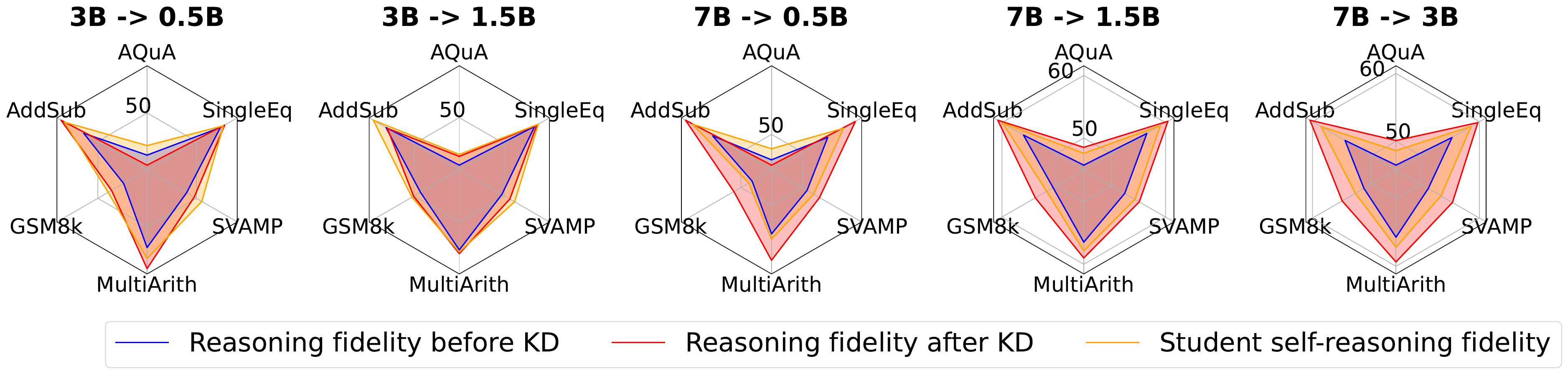}\label{fig:student_fidelty}}
    \quad
    \subfloat[Instruction following tasks]{\includegraphics[width=0.88\linewidth]{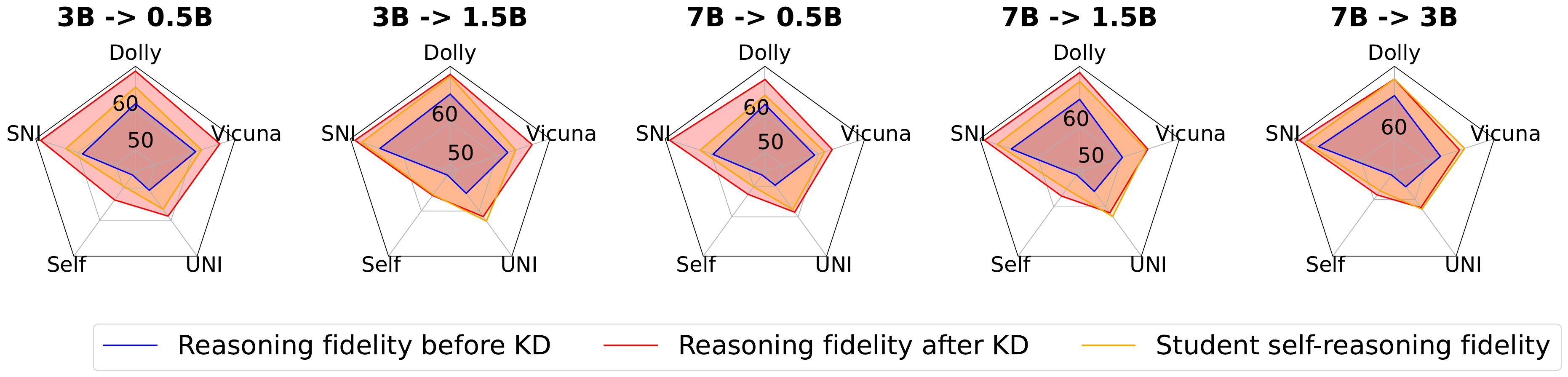}\label{fig:student_fidelty_int}}
    \caption{Student reasoning fidelity with teacher model on (a) mathematical reasoning and (b) instruction following tasks. We omit the commonsense reasoning tasks for this analysis as the reasoning steps are not generated. Results with Qwen-14B and LLaMA-8B teacher are highlighted in Figures~\ref{fig:student_fidelty2} and~\ref{fig:student_fidelty_llama} of Appendix~\ref{appx:results}, respectively.}
    
\end{figure*}
\paragraph{How does KD impair teacher-student agreement?}  
Figure~\ref{fig:student_agreement} and Figure~\ref{fig:student_agreement2} of Appendix~\ref{appx:results} show that post-KD agreement between student and teacher models declines as student size increases. Smaller models (e.g., Qwen-0.5B and Qwen-1.5B) exhibit higher agreement with larger Qwen-14B model in structured mathematical tasks like AddSub ($89.1\%$) and MultiArith ($94.5\%$), indicating effective transfer of well-defined rules. 

\begin{table*}[htbp]
  \centering
  \subfloat[Mathematical reasoning tasks]{\scalebox{0.8}{
    \begin{tabular}{l|llllll}
    \cline{1-7}
    \textbf{Test Type} & \textbf{AQuA} & \textbf{AddSub} & \textbf{GSM8k} & \textbf{MultiArith} & \textbf{SVAMP} & \textbf{SingleEq} \\
    \cline{1-7}
    Spearman rank & -0.09 (0.6) & 0.01 (1.0) & 0.01 (0.9) & -0.25 (0.1) & -0.07 (0.7) & -0.12 (0.5)\\
    \cline{1-7}
    \end{tabular}%
    }}
    \quad
    \subfloat[Instruction following tasks]{\scalebox{0.8}{
    \begin{tabular}{l|lllll}
    \cline{1-6}
    \textbf{Test Type} & \textbf{Dolly} & \textbf{Self} & \textbf{SNI} & \textbf{UNI} & \textbf{Vicuna} \\
    \cline{1-6}
    Spearman rank & 0.3 (0.2) & -0.06 (0.8) & -0.06 (0.8) & -0.13 (0.6) & 0.07 (0.8) \\
    \cline{1-6}
    \end{tabular}%
    }}
      \caption{Spearman rank correlation and $p$-value between student performance and teacher-student fidelity.} 
  \label{tab:hypo5}%
  \vspace{-5mm}
\end{table*}%

Table~\ref{tab:hypo4math} reveals that higher agreement does not always translate to better student performance (e.g., negative correlation for MultiArith and SVAMP) suggesting that effective students deviate from teacher outputs rather than mimicking them.
For commonsense reasoning (reported in Table~\ref{tab:hypo4cs}), the correlation between teacher-student agreement and the student performance remains weak, indicating statistically insignificant correlation. 

\begin{figure*}[!t]
    \centering
    \subfloat[Mathematical reasoning]{\includegraphics[width=0.41\linewidth]{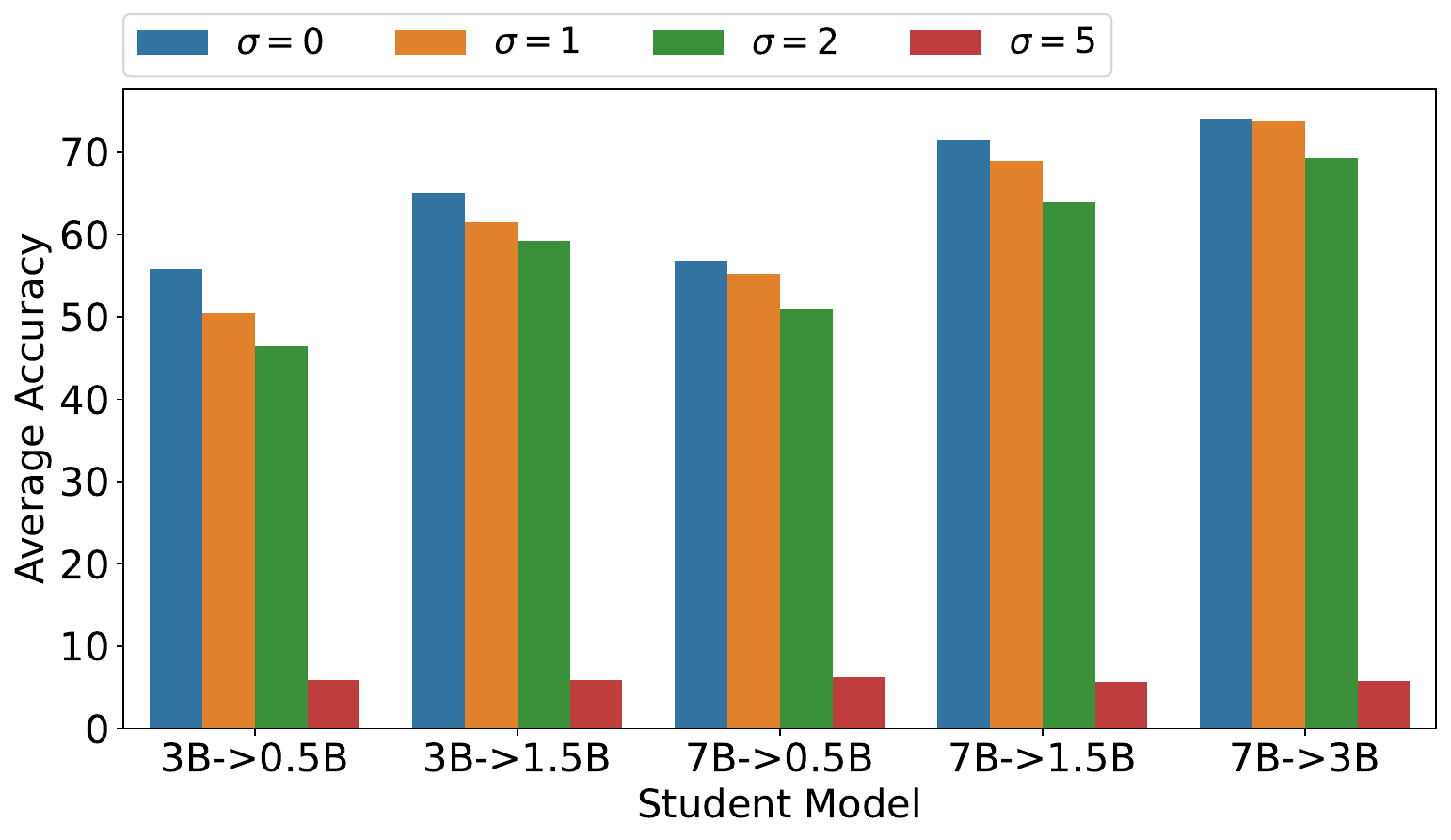}}
    \subfloat[Commonsense reasoning]{\includegraphics[width=0.41\linewidth]{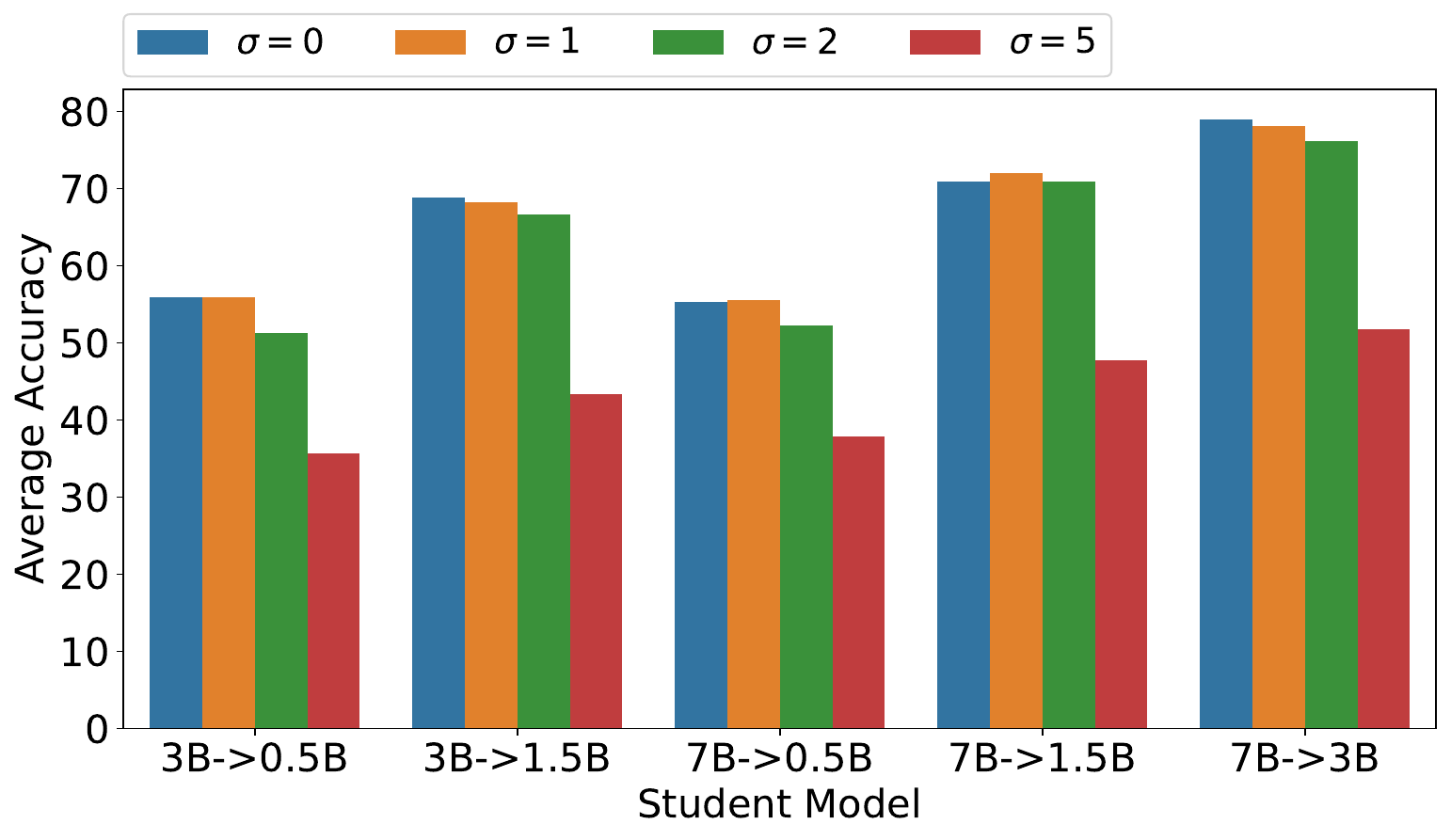}}
    \caption{Impact of teacher input noise scale ($\sigma$) on student's downstream performance on (a) mathematical and (b) commonsense reasoning tasks.}
    \label{fig:noise_ablation}
    \vspace{-5mm}
\end{figure*}

\begin{figure*}[!t]
    \centering
    \subfloat[Mathematical reasoning]{\includegraphics[width=0.41\linewidth]{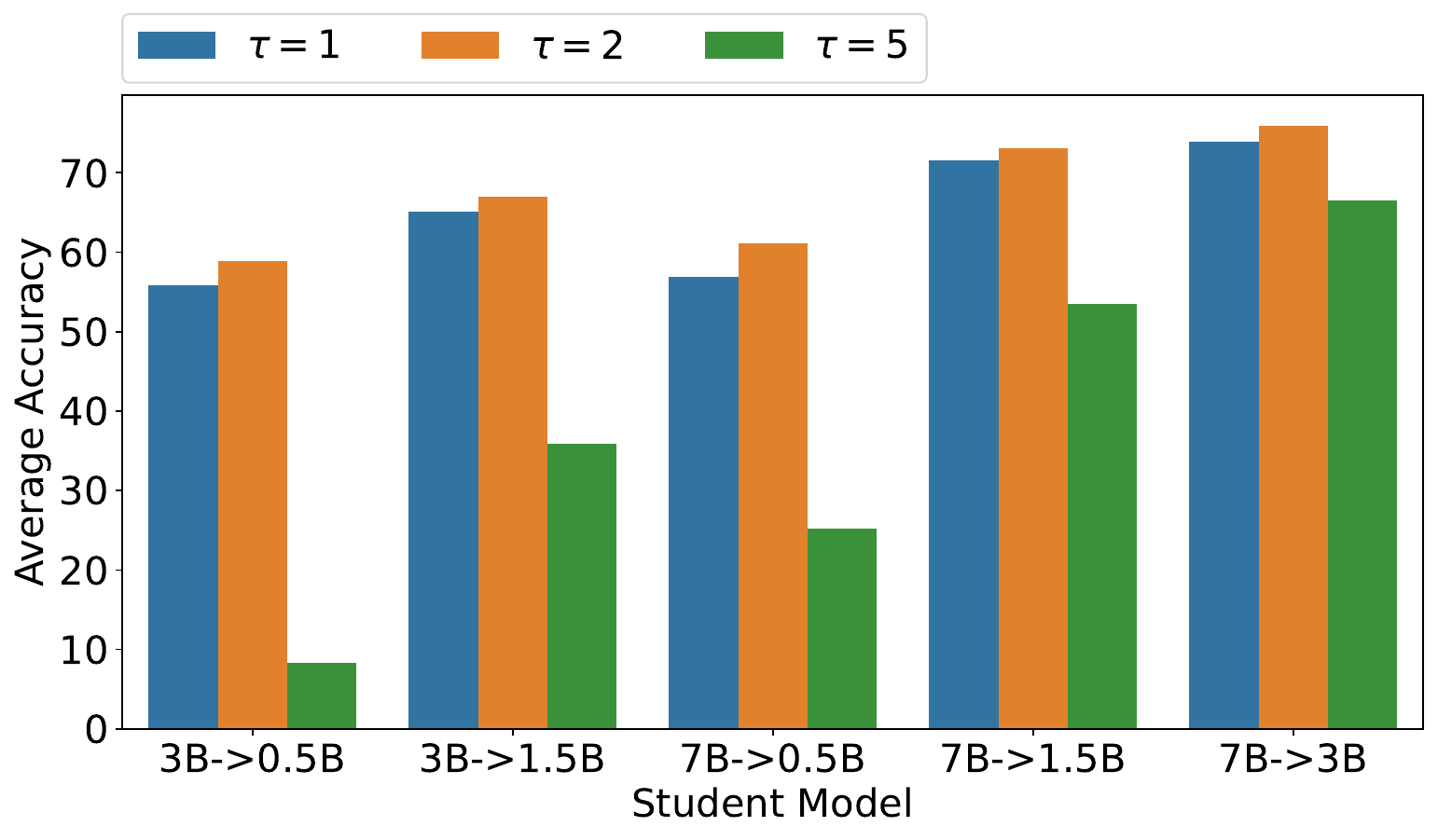}}
    \subfloat[Commonsense reasoning]{\includegraphics[width=0.41\linewidth]{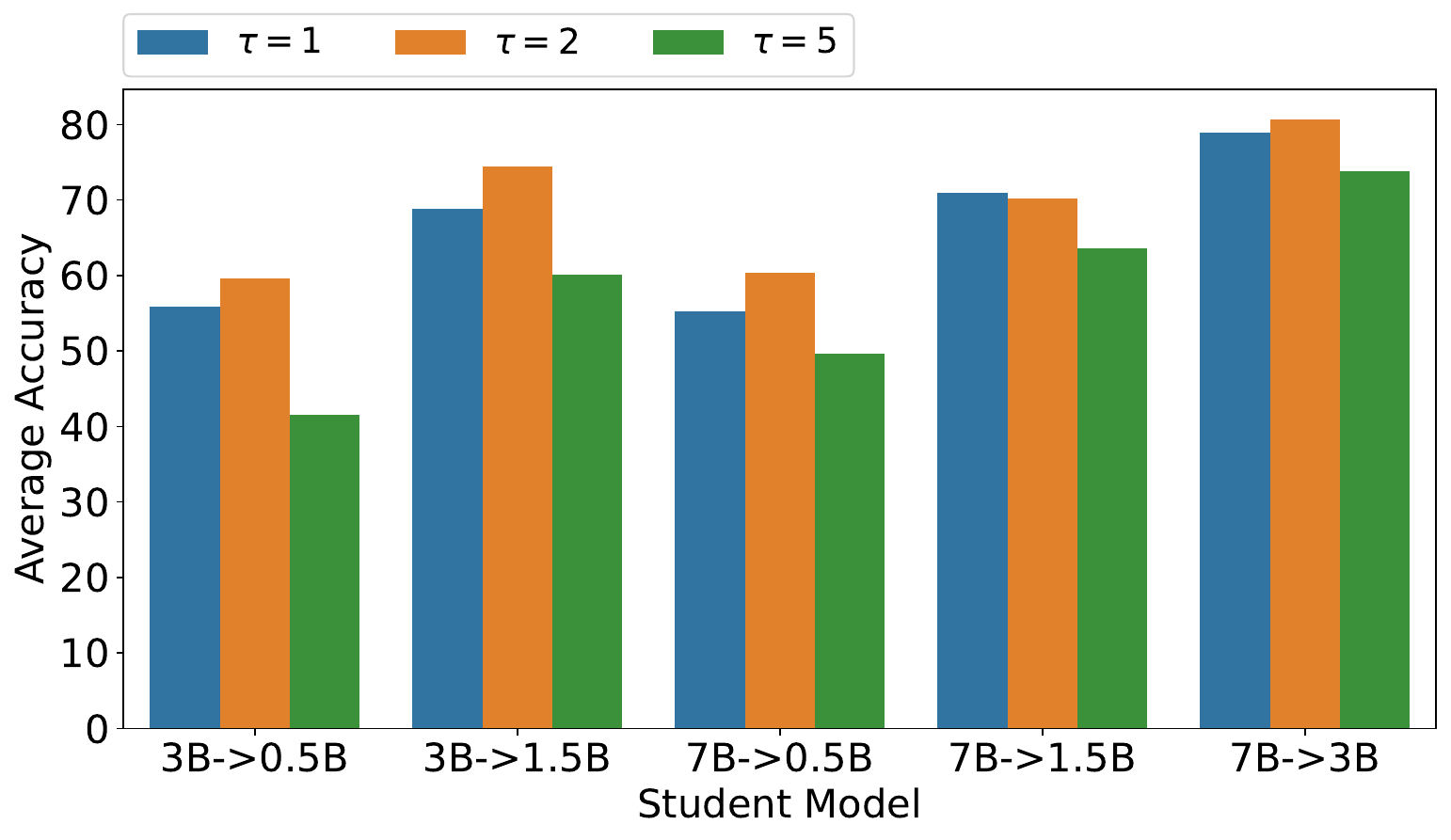}}
    \caption{Impact of smoothing factor ($\tau$) on student's downstream performance on mathematical (a) and commonsense (b) reasoning tasks.}
    \label{fig:temp_ablation}
    \vspace{-5mm}
\end{figure*}

\paragraph{How does KD impair teacher-student fidelity?}  
Teacher-student fidelity, measuring alignment in reasoning patterns, varies by model size and distillation method. Along with the BLEU-based fidelity metric, we devise an alternative fidelity metric, calculated using the cosine similarity between sentence embeddings~\citep{reimers-2019-sentence-bert} of teacher and student responses. The Spearman correlation between BLEU scores and cosine similarity is 0.97 (p-value = 1e-9), indicating strong agreement. Therefore, we report only BLEU-based fidelity metric, as the resulting hypotheses remain unchanged. While KD improves fidelity over SFT (Figure~\ref{fig:student_fidelty}), Table~\ref{tab:hypo5} indicates weak correlations between fidelity and student performance, implying that strict imitation does not necessarily enhance performance. For instance, AddSub ($0.01$, $p$-value=$1.0$) and GSM8k ($0.01$, $p$-value=$0.9$) show low correlations, reinforcing that high fidelity alone is not a reliable predictor of KD effectiveness. Error analysis in Table~\ref{tab:error_analysis3} of Appendix~\ref{appx:error_analysis} further highlights the feeble connection between teacher-student fidelity and student generalization.

\paragraph{Impact of noisy teacher signals on KD.}  
Considering the low fidelity of KD despite its improvements in generalization, we investigate the significance of teacher signals by injecting Gaussian noise (\(\epsilon \sim \mathcal{N}(0, \sigma)\)) into teacher logits before distillation. Figure~\ref{fig:noise_ablation} shows that increasing \(\sigma\) from 0 (no noise) to 1 slightly reduces performance (e.g., Qwen-3B $\rightarrow$ Qwen-1.5B drops by $3.6\%$). A further increase to \(\sigma=2\) continues the decline (e.g., Qwen-7B $\rightarrow$ Qwen-3B falls by $4.7\%$). At \(\sigma=5\), performance collapses, with Qwen-3B $\rightarrow$ Qwen-0.5B plummeting from $55.87\%$ to $5.85\%$, confirming that excessive noise disrupts knowledge transfer. Commonsense reasoning follows a similar trend but is slightly more robust, with Qwen-3B $\rightarrow$ Qwen-1.5B maintaining $66.62\%$ at \(\sigma=2\) compared to its mathematical counterpart ($59.26\%$). However, at \(\sigma=5\), performance sharply declines (e.g., Qwen-7B $\rightarrow$ Qwen-3B drops from $78.94\%$ to $51.76\%$), reinforcing that moderate noise may aid generalization, but excessive noise severely impairs KD.

\paragraph{Impact of temperature smoothing.}  
Figure~\ref{fig:temp_ablation} highlights that temperature (\(\tau\)) significantly impacts KD effectiveness. A moderate \(\tau=2\) consistently yields the best results across both reasoning types. In mathematical reasoning tasks, increasing \(\tau\) from 1 to 2 improves all models (e.g., Qwen-3B $\rightarrow$ Qwen-1.5B rises from $65.09\%$ to $66.93\%$, Qwen-7B $\rightarrow$ Qwen-3B reaches $75.95\%$). However, \(\tau=5\) leads to severe performance drops, especially for smaller students (e.g., Qwen-3B $\rightarrow$ Qwen-0.5B drops to $8.3\%$), indicating that excessive smoothing weakens the learning signal. Larger students tolerate higher temperatures better (e.g., Qwen-7B $\rightarrow$ Qwen-3B retains $66.51\%$ at \(\tau=5\)). Commonsense reasoning follows a similar pattern, but performance degradation at \(\tau=5\) is less severe compared to mathematical tasks. These findings emphasize that temperature tuning is crucial, with optimal \(\tau\) values varying by student size and task complexity.

\paragraph{Impact of teacher-student gap on KD.} 
~\citet{Mirzadeh_Farajtabar_Li_Levine_Matsukawa_Ghasemzadeh_2020} observe that in computer vision models, student performance tends to degrade when the gap between the teacher and student is too large. We find a similar trend in our setting: the most effective teacher for a student model is not necessarily the largest one available. For instance, as shown in Table 7, Qwen-1.5B achieves its best performance when distilled from Qwen-7B rather than the larger Qwen-14B. We observe this pattern across multiple knowledge distillation methods and datasets.

\section{Conclusion}
This paper elaborated the impact of  KD on small LMs, considering factors like teacher performance, student size, and distillation methods across mathematical and commonsense reasoning tasks. Results showed that KD significantly benefits smaller models, but its effectiveness diminishes with increasing model size. Teacher domain adaptation played a more critical role than teacher performance in KD success, particularly for structured reasoning tasks. Surprisingly, higher teacher-student agreement did not always correlate with better student performance, especially in complex reasoning tasks where strong students often deviated from teacher outputs. These findings underscore the need for task-aware KD strategies and adaptive distillation techniques tailored to student learning dynamics. Future research should explore alternative KD objectives, self-distillation mechanisms, and refined teacher-student alignment strategies to improve both performance and reasoning fidelity. 

\newpage

\section*{Limitation}
While this study provides a comprehensive evaluation of KD across diverse reasoning tasks, certain aspects remain open for further exploration. Firstly, our experiments focus on a select set of reasoning tasks, and while the findings generalize well within these domains, future work could extend the analysis to broader task distributions, including multimodal learning and domain-specific applications. Secondly, while we investigate multiple KD techniques, our study primarily evaluates teacher-student distillation in a single-step process; iterative and multi-teacher KD frameworks may further enhance performance and warrant deeper investigation.

\section*{Ethical Considerations}
We acknowledge the potential ethical concerns associated with knowledge distillation, such as the risk of bias propagation from teacher models to students and the possible loss of interpretability in distilled models. Our analysis highlights cases where KD enhances accuracy but does not always preserve reasoning fidelity, raising concerns about trustworthiness in critical applications. To mitigate these risks, we encourage the development of more interpretable KD techniques and stress the importance of evaluating distilled models not only for performance but also for fairness, robustness, and alignment with human reasoning.

\section*{Acknowledgment}
We acknowledge the support of
the IBM-IITD AI Horizons network. T. Chakraborty acknowledges the support of the Rajiv
Khemani Young Faculty Chair Professorship in Artificial Intelligence.

\bibliography{custom}

\appendix

\section{Datasets} \label{appx:datasets}
The Mathematical Reasoning benchmark consists of the following datasets:
\begin{itemize}
\itemsep0em
\item GSM8K(Grade School Math 8k)~\cite{cobbe2021training} - This dataset consists of basic math problems from grade school that require multi-step reasoning.

\item SVAMP(Simple Variations on Arithmetic Math word Problems)~\cite{patel2021nlp} - This is a challenge set that tests a model across different aspects of Math Word Problems like testing whether a model is Question sensitive,  has robust reasoning ability or is invariant to structural alterations in questions.

\item MultiArith~\cite{roy2015solving} - This dataset consists of arithmetic problems with multiple steps and basic mathematical operations.

\item AddSub~\cite{hosseini2014learning} - This dataset consists of arithmetic problems involving just addition and subtraction.

\item AQuA~\cite{ling2017program} - This dataset consists of algebraic word problems with answer rationales.

\item SingleEq~\cite{koncel2015parsing} - This dataset contains math word problems that can be expressed in a singl equation.

\item Math10K~\cite{hu2023llm} - This dataset contains training examples from GSM8K, AQuA, MAWPS and MAWPS-single~\cite{koncel-kedziorski-etal-2016-mawps}. The original training examples only contain equations and final answers. Hence the authors used ChatGPT to generate intermediate reasoning steps for each training example to curate the final Math10K dataset.

\end{itemize}

The Commonsense reasoning benchmark consists of the following datasets:
\begin{itemize}
\itemsep0em
\item Hellaswag~\cite{zellers2019hellaswag} - This dataset is used for evaluating commonsense NLI. Authors use Adversial Filtering to select a challenging set of examples.

\item Winogrande~\cite{sakaguchi2021winogrande} - This dataset contains fill-in the blank problems that are inspired by the original Winograd Schema Challenge, but modified to improve scale and robustness against dataset-specific bias. Given two options, the goal is to choose the right option for a given sentence which requires commonsense reasoning.

\item ARC~\cite{allenai:arc} - This dataset contains multiple-choice question answers from grade school science exams. The dataset is split into "Challenge set" and "Easy set", with the "Challenge" set including only those examples that were incorrectly answered by both a retrieval-based algorithm and a word co-occurrence algorithm.

\item OBQA (Open Book Question Answering)~\cite{OpenBookQA2018} - This is a new kind of question-answer dataset that is modeled after open-book exams. It contains questions that need multi-step reasoning and broad common knowledge to answer them.

\item BoolQ~\cite{clark2019boolq} - This dataset contains questions that can be answered with either a yes or no as the answer. The questions are gathered from queries to the google search engine. They are filtered and annotated by humans.

\item PiQA (Physical Interaction Question Answering)~\cite{Bisk2020} - This dataset introduces the task of physical commonsense reasoning to investigate physical knowledge of models. 

\item SiQA (Social Interaction Question Answering)~\cite{sap-etal-2019-social} - This dataset is used for testing social commonsense intelligence. It contains questions related to a wide variety of social interactions. Answer options include both human curated answers and machine generated answers.

\item Commonsense-15K~\cite{hu2023llm} - This dataset contains examples from BoolQ, PiQA, SiQA, Hellaswag, Winogrande, ARC-e, ARC-c and OBQA. Authors use a structured template by first describing the particular task's goal, followed by the content and answer of the example. 
\end{itemize}

The Instruction following benchmark consists of the following datasets:
\begin{itemize}
\itemsep0em
\item Dolly~\cite{gu2024minillm} - Following ~\citet{gu2024minillm}, we use a filtered set from databricks-dolly-15K containing about 12.5k samples for training, 1k samples for validation and 500 samples for testing.

\item Self~\cite{selfinstruct} - This dataset consists of 252 user-oriented instruction-following samples.

\item Vicuna~\cite{vicuna2023} - This dataset contains 80 challenging questions synthesized by GPT-4 used in Vicuna evaluation.

\item SNI~\cite{wang-etal-2022-super} - The dataset comprises approximately 9K samples drawn from around 119 tasks within the Super-Natural Instructions benchmark. Following ~\citet{gu2024minillm}, we divide the samples into three subsets based on the length of the ground-truth responses. For our experiments, we use the subset with response lengths in the range [11, $\infty$] which contains about 1.6K samples.

\item UNI~\cite{honovich-etal-2023-unnatural} - This dataset consists of samples from the core set of Unnatural Instructions. As with S-NI, we focus on the subset where ground-truth response lengths fall within the range [11, $\infty$]. We use first 2k samples of them for evaluation.

\end{itemize}

The train/val/test dataset splits for mathematical, commonsense reasoning and instruction following datasets are highlighted in Table~\ref{tab:data_split_float}.

\newcommand{\mathsplit}
{
    \begin{tabular}{lccc}
\toprule
     Dataset     & \# train & \# validation & \# test \\ \midrule
         Math10k    & 10K       & 500             & -       \\ 
         GSM8K       & 8.8K     & -             & 1319    \\ 
         SVAMP       & -        & -             & 1000    \\ 
         MultiArith  & -        & -             & 600     \\ 
         AddSub      & -        & -             & 395     \\ 
         AQuA        & 1K       & -             & 254     \\ 
         SingleEq    & -        & -             & 508     \\ 
\bottomrule
\end{tabular}
}

\newcommand{\cssplit}
{
    \begin{tabular}{lccc}
\toprule
     Dataset     & \# train & \# validation & \# test \\ \midrule
         Commonsense15K    & 15K      & 500             & -       \\ 
         Hellaswag       & 3.5K     & -             & 10K    \\ 
         Winogrande       & 5.5K        & -             & 1.2K    \\ 
         ARC-c  &  100        & -             & 1.1K     \\ 
         ARC-e      & 200        & -             & 2.3K     \\ 
         OBQA & 500  & - & 500 \\
         BoolQ & 800 & - & 3.2K \\
         PiQA        & 1.5K     & -             & 2K     \\ 
         SiQA    &  3K        & -            & 2K     \\ 
\bottomrule
\end{tabular}
}

\newcommand{\ifsplit}
{
    \begin{tabular}{lccc}
\toprule
     Dataset     & \# train & \# validation & \# test \\ \midrule
         Dolly15K    & 12.5K       & 1k         & 500   \\ 
         SelfInst    & -     & -                & 252   \\ 
         VicunaEval  & -        & -            & 80    \\ 
         S-NI  & -        & -             & 1.6K     \\ 
         UnNI      & -        & -             & 2k     \\  
\bottomrule
\end{tabular}
}

\begin{table}[!htb]
  \centering
    \subfloat[Mathematical reasoning tasks]{\scalebox{0.98}\mathsplit
    \label{tab:math_split}}
    \quad
    \subfloat[Commonsense reasoning tasks]{\scalebox{0.85}\cssplit
    \label{tab:cs_split}}
    \quad
    \subfloat[Instruction following tasks]{\scalebox{0.9}\ifsplit
    \label{tab:if_split}}
    \caption{Dataset splits for all tasks.} 
  \label{tab:data_split_float}%
\end{table}

\begin{figure*}[!htb]
    \centering
    \subfloat[Performance on mathematical reasoning tasks]{
    \includegraphics[width=0.45\linewidth]{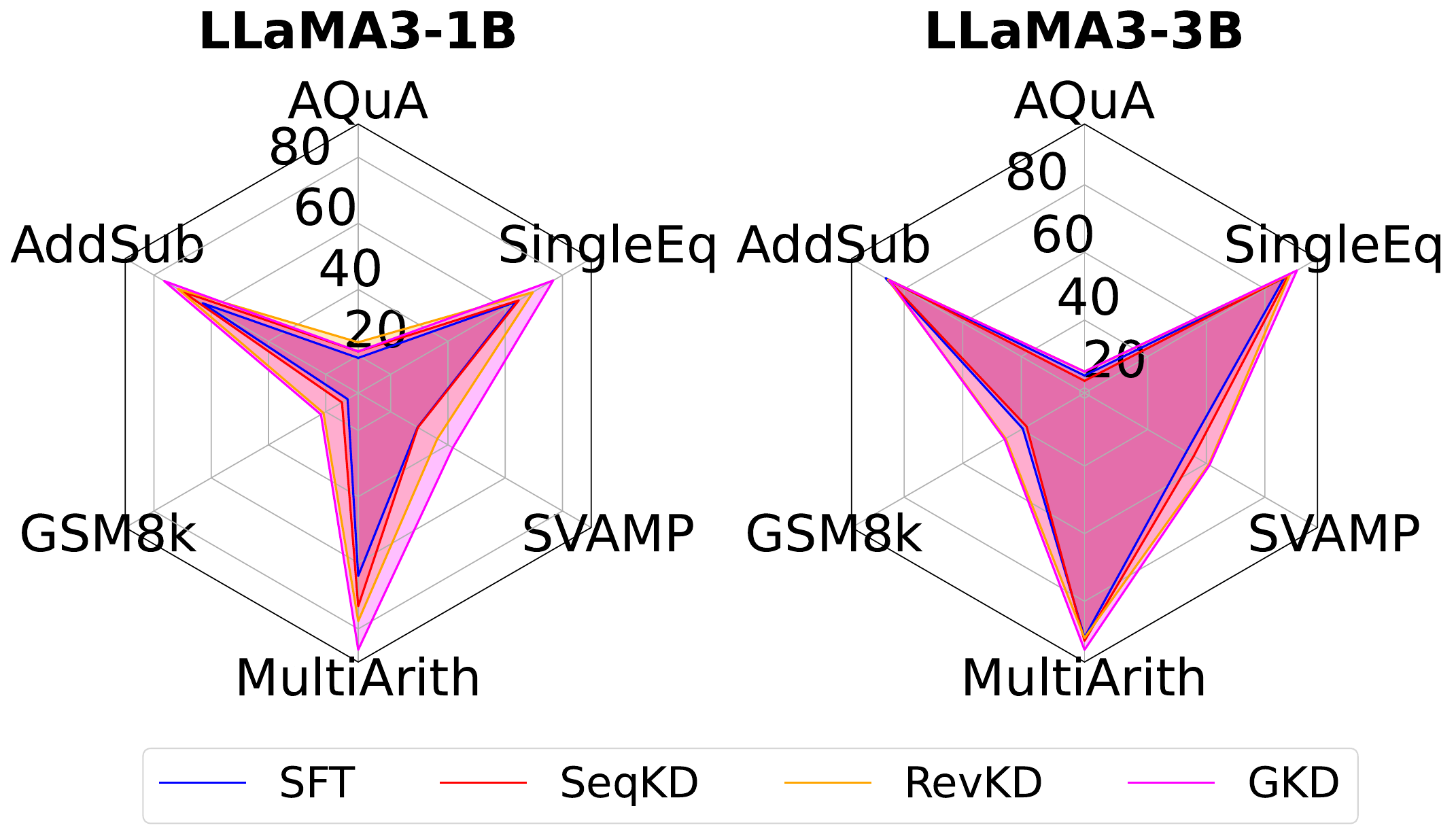}}
    \subfloat[Performance on commonsense reasoning tasks]{
    \includegraphics[width=0.5\linewidth]{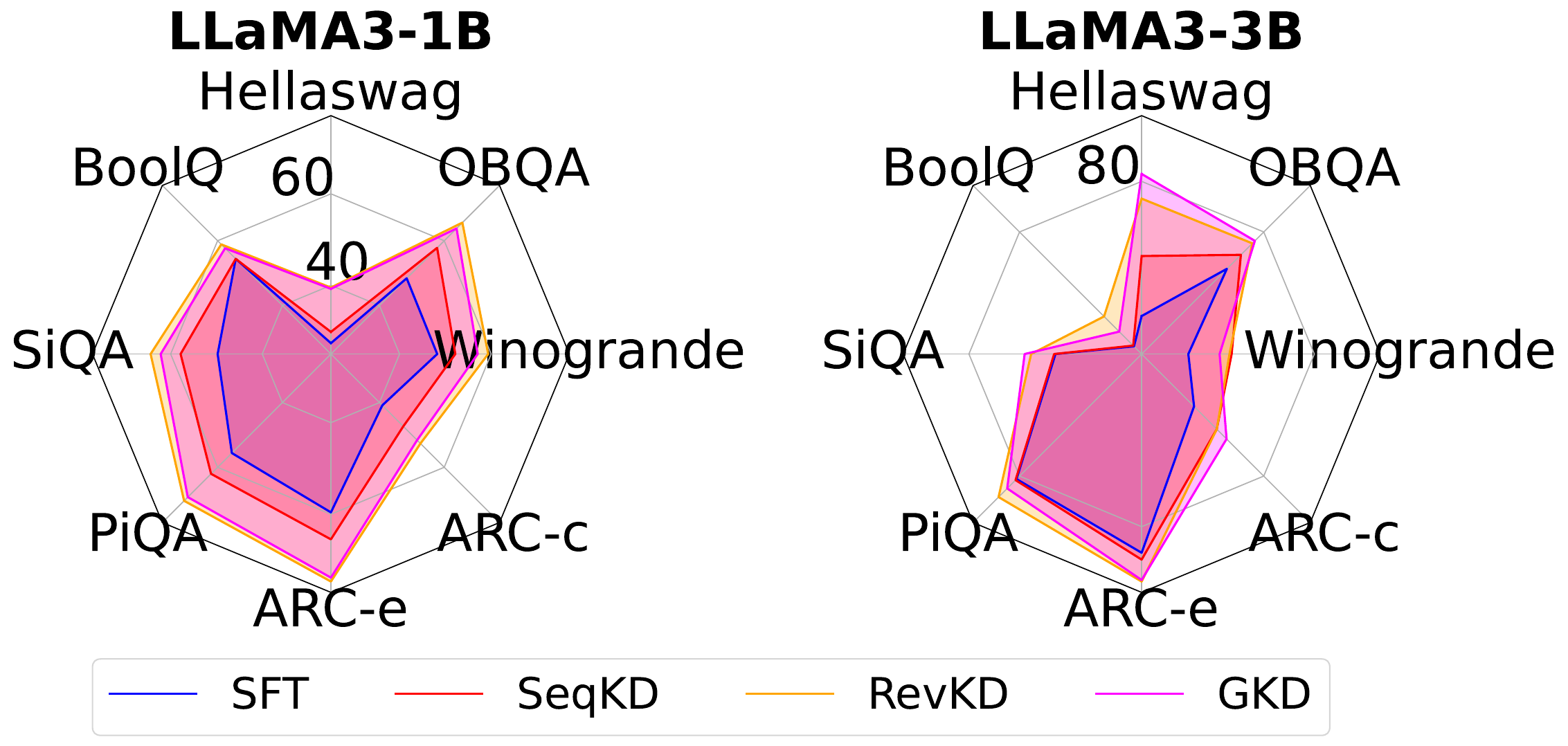}}
    \quad
    \subfloat[Performance on instruction following tasks]{
    \includegraphics[width=0.5\linewidth]{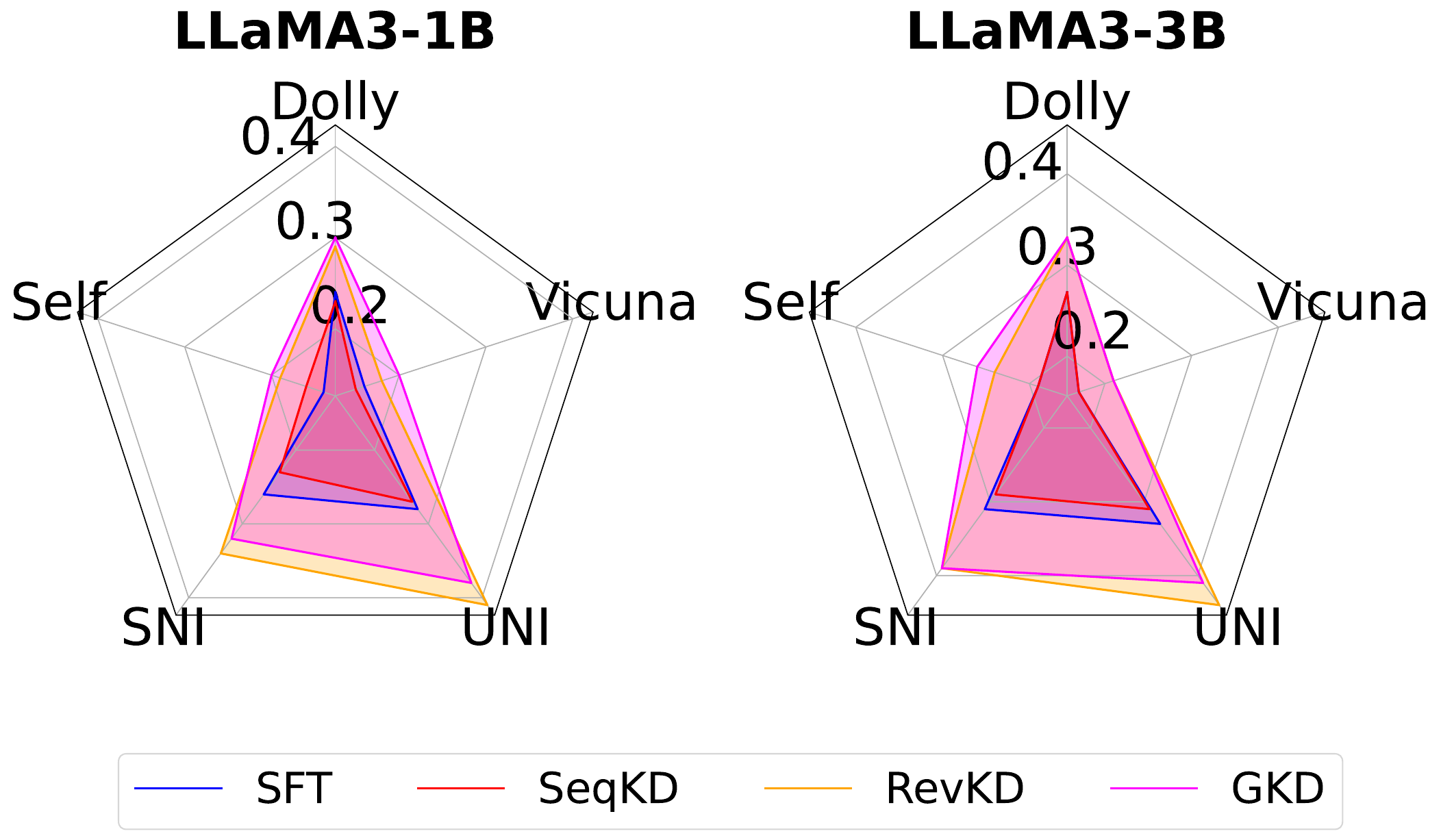}}
    \caption{Performance of LLaMA-3 student models on different mathematical reasoning (a) commonsense reasoning (b) and instruction following (c) tasks without and with distillation from LLaMA-3-8B model.}
    \label{fig:student_performance_llama}
\end{figure*}

\begin{figure*}[!htb]
    \centering
    \subfloat[Mathematical reasoning tasks]{
    \includegraphics[width=0.95\linewidth]{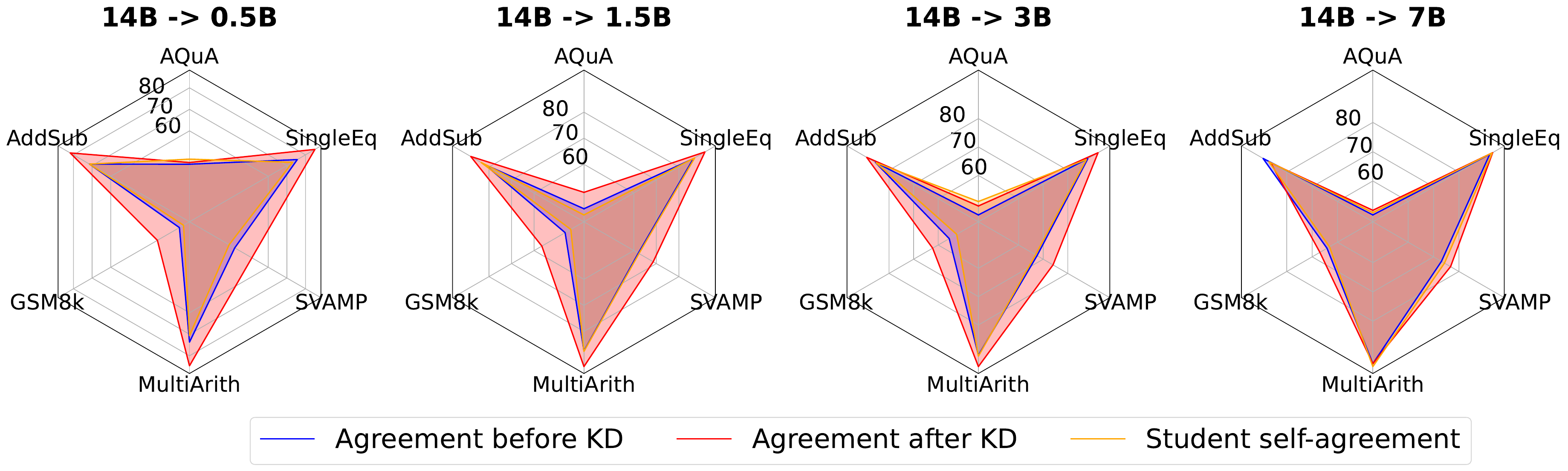}}
    \quad
    \subfloat[Commonsense reasoning tasks]{\includegraphics[width=0.95\linewidth]{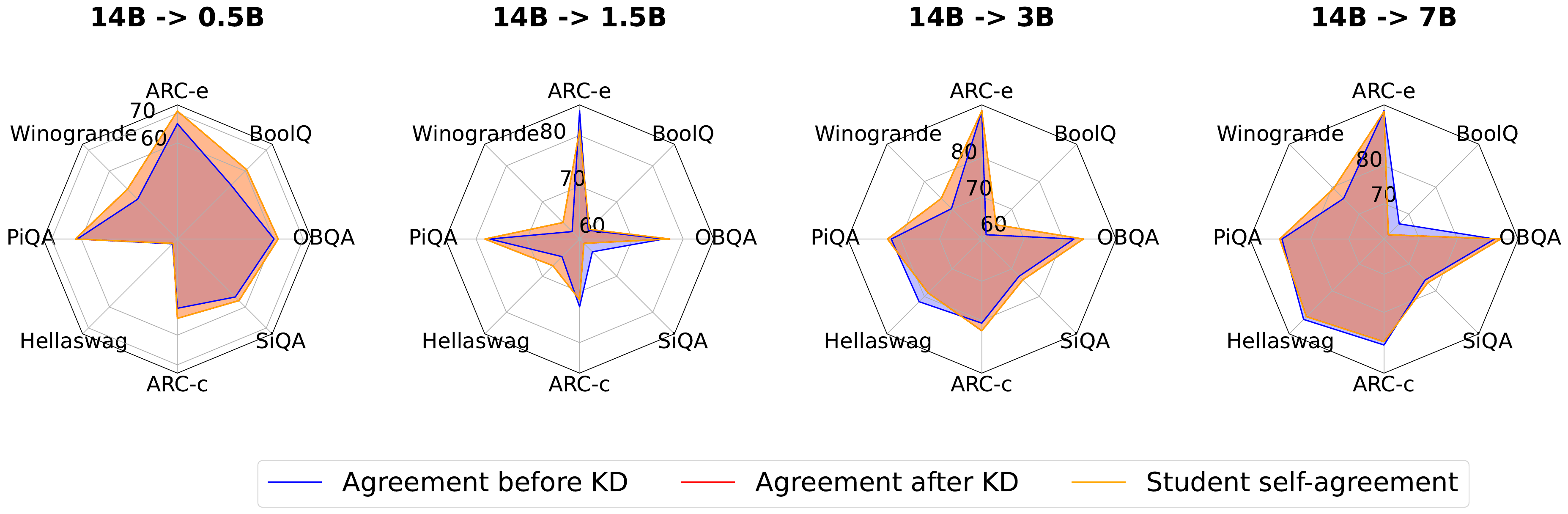}
    }
    \caption{Student agreement with Qwen-14B teacher.}
    \label{fig:student_agreement2}
\end{figure*}

\begin{figure*}[!htb]
    \centering
    \subfloat[Mathematical reasoning tasks]{\includegraphics[width=0.95\linewidth]{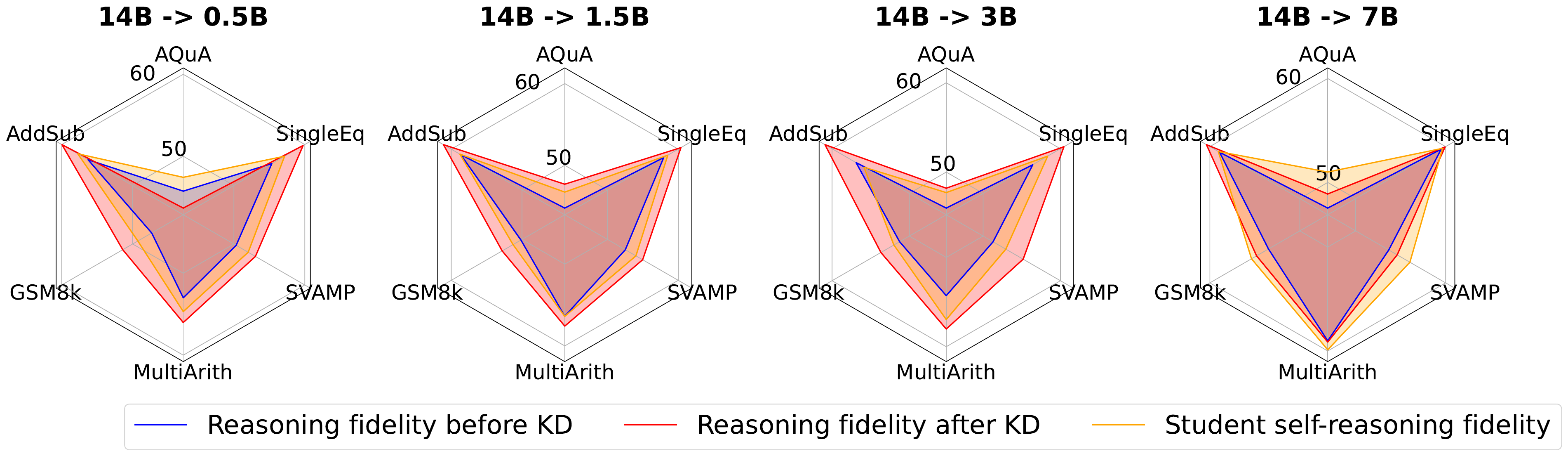}
    }
    \quad
    \subfloat[Instruction following tasks]{\includegraphics[width=0.95\linewidth]{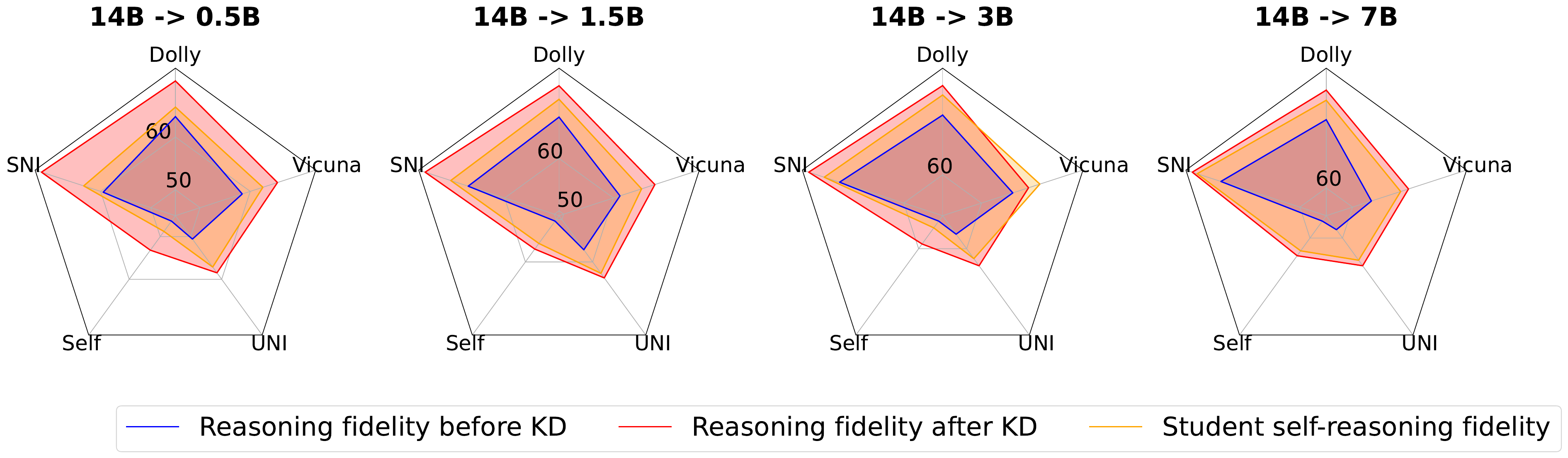}\label{fig:student_fidelty_int2}}
    \caption{Student reasoning fidelity for Qwen-14B teacher model.}
    \label{fig:student_fidelty2}
\end{figure*}

\begin{figure*}[!htb]
    \centering
    \subfloat[Mathematical reasoning tasks]{
    \includegraphics[width=0.5\linewidth]{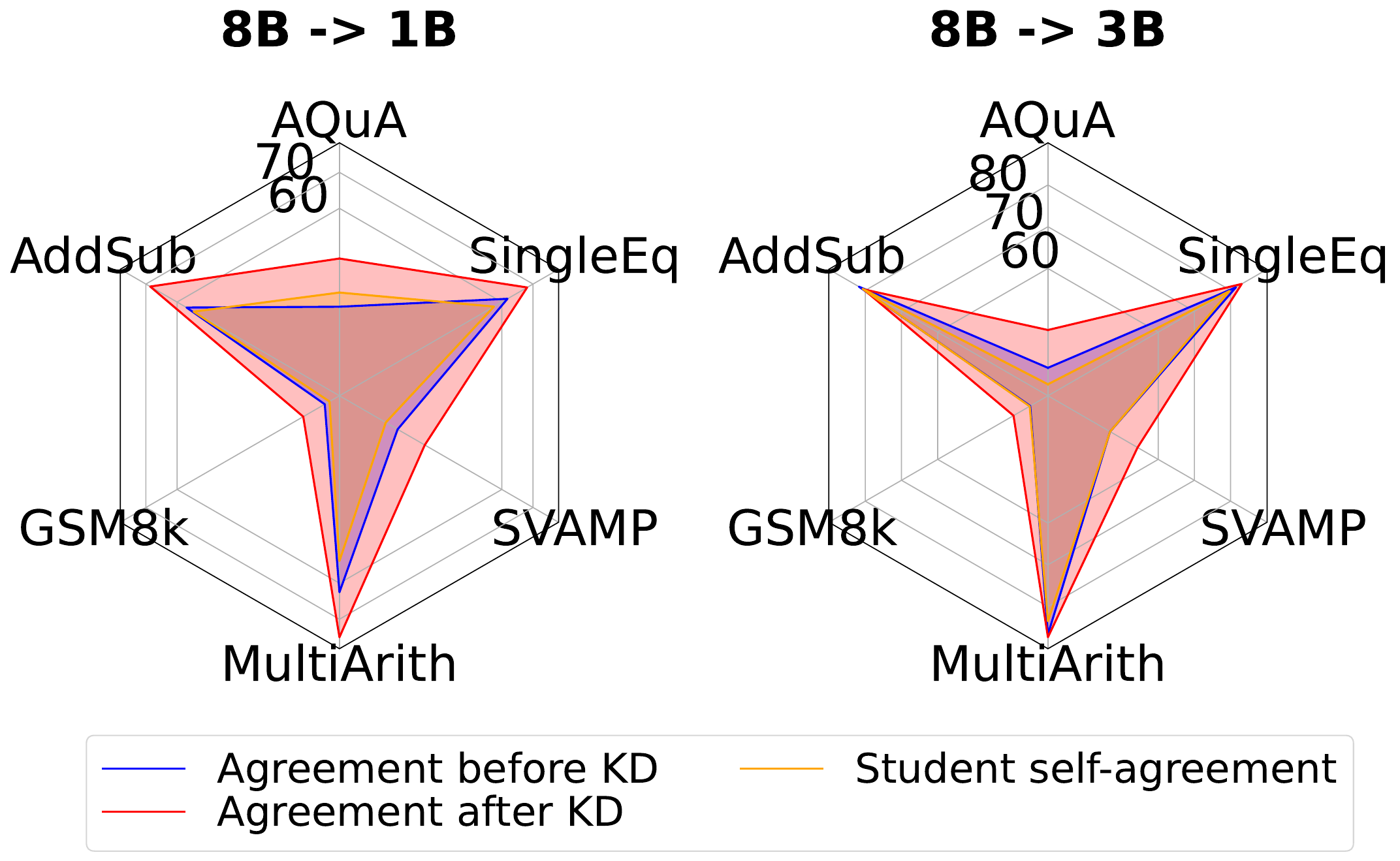}}
    \subfloat[Commonsense reasoning tasks]{\includegraphics[width=0.5\linewidth]{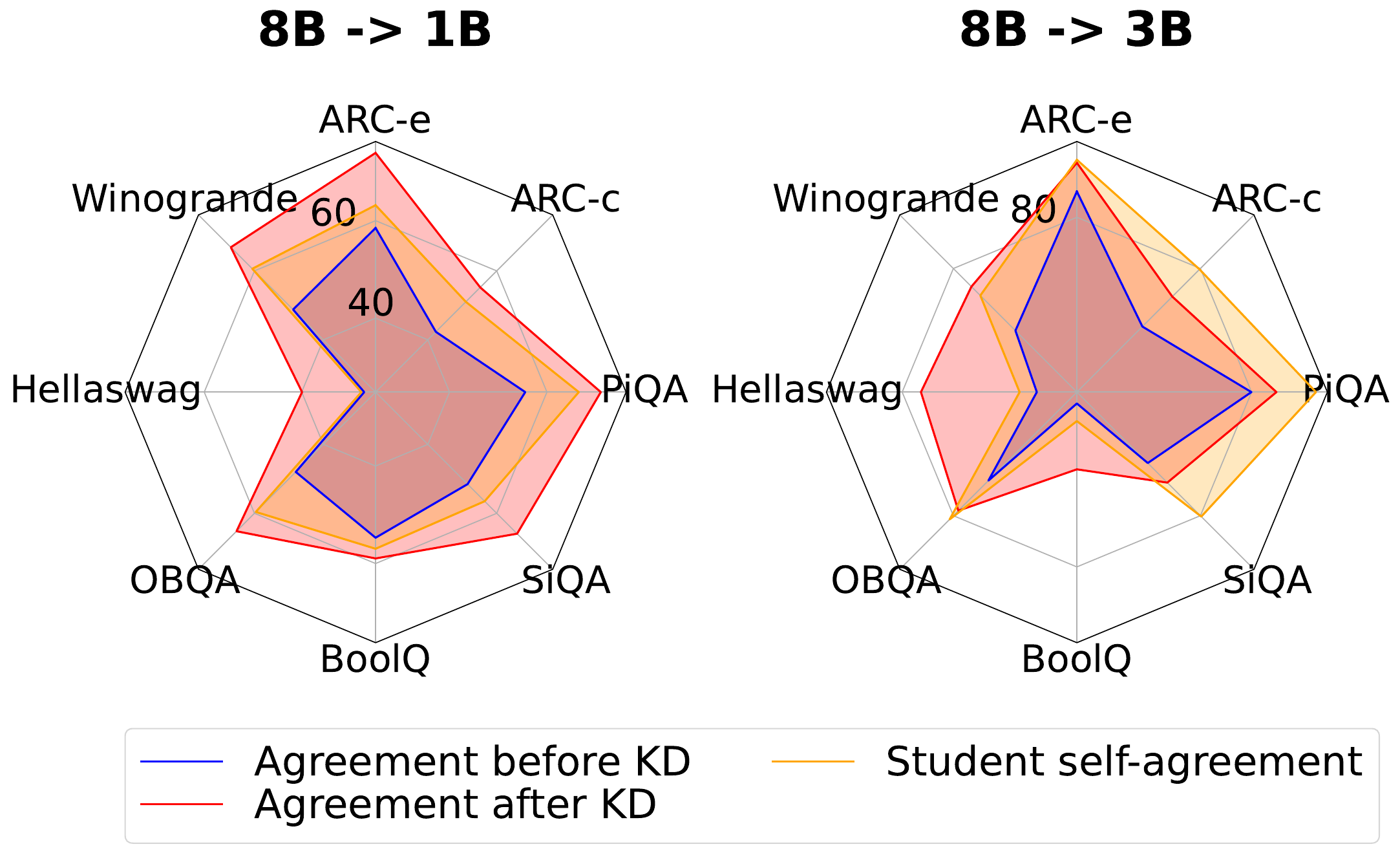}
    }
    \caption{Student agreement with LLaMA-3-8B teacher.}
    \label{fig:student_agreement_llama}
\end{figure*}

\begin{figure*}[!htb]
    \centering
    \subfloat[Mathematical reasoning tasks]{\includegraphics[width=0.7\linewidth]{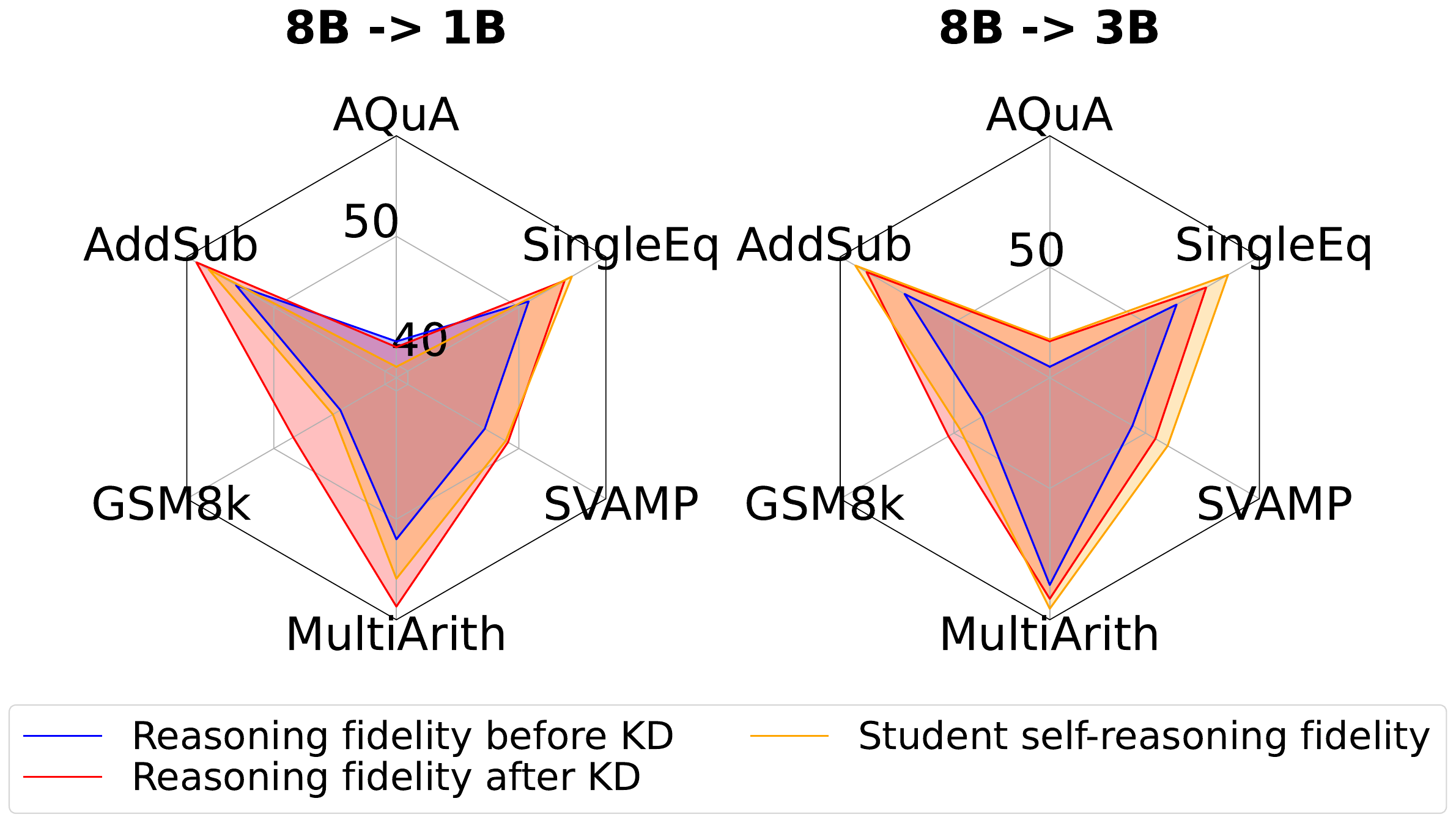}
    }
    \quad
    \subfloat[Instruction following tasks]{\includegraphics[width=0.7\linewidth]{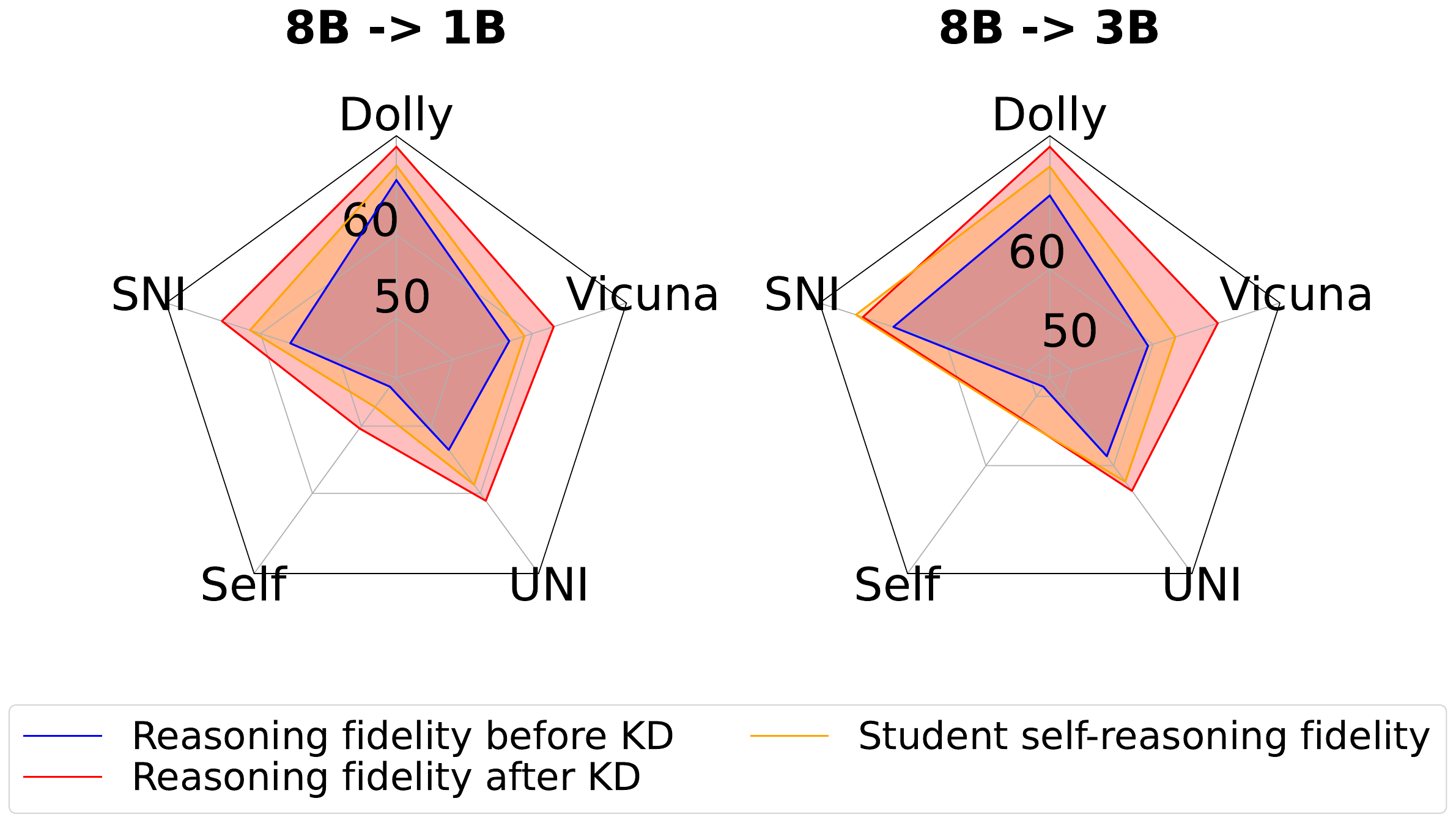}
    \label{fig:student_fidelty_llama_int}}
    \caption{Student reasoning fidelity for LLaMA-3-8B teacher model.}
    \label{fig:student_fidelty_llama}
\end{figure*}

\section{Results}\label{appx:results}
\paragraph{Impact of KD on student generalization.} We report the detailed results for SFT, SeqKD, RevKD and GKD of all teacher-student combinations for Math Reasoning tasks in Table~\ref{tab:math_all_results}, Commonsense reasoning tasks in Table~\ref{tab:cs_all_results} and Instruction following tasks in Table~\ref{tab:dolly_all_results}. RevKD consistently delivers the best performance, with Qwen-3B distilled from Qwen-7B achieving $76.91\%$, outperforming SeqKD ($73.96\%$) and GKD ($76.18\%$). However, larger student models benefit less from KD, as seen in Qwen-7B distilled from Qwen-14B, which only shows a marginal improvement over its fine-tuned counterpart ($77.81\%$ $\rightarrow$ $78.02\%$). Structured tasks such as MultiArith and SingleEq exhibit the highest gains, often exceeding $90\%$ accuracy, indicating that KD effectively transfers arithmetic-based reasoning. Conversely, AQuA and GSM8K remain more challenging, particularly for smaller students, highlighting that complex multi-step reasoning is harder to distill. Additionally, the results suggest that KD effectiveness does not depend on model architecture, as LLaMA-based students benefit similarly from KD, with LLaMA-1B distilled from LLaMA-8B with GKD achieving $54.03\%$ imporving the SFT model by $12\%$. Finally, distillation from stronger teachers improves student performance, but not linearly, as Qwen-3B distilled from Qwen-7B performs almost as well as when distilled from Qwen-14B, reinforcing that teacher expertise matters more than sheer size. Overall, these findings underscore that KD significantly enhances smaller models' reasoning capabilities, but diminishing returns appear for larger models, necessitating adaptive KD strategies tailored to student capacity and task complexity.

On commonsense reasoning tasks also RevKD generally outperforms other KD methods, with Qwen-7B distilled from Qwen-14B achieving $85.26\%$, surpassing both SeqKD ($83.55\%$) and GKD ($84.90\%$). Similarly, Qwen-3B distilled from Qwen-14B using RevKD reaches $80.79\%$, showing a consistent advantage over SeqKD ($78.77\%$) and GKD ($80.38\%$). Larger models like Qwen-7B exhibit only marginal gains post-distillation, indicating that they already possess strong commonsense reasoning capabilities. Structured tasks such as ARC-e, ARC-c, and PiQA benefit most from KD, as they involve multiple-choice reasoning, where teacher guidance is transferred effectively. However, open-ended tasks such as BoolQ and SiQA show smaller improvements, suggesting that KD is less effective in distilling nuanced, context-dependent reasoning. 
Additionally, LLaMA-based models benefit similarly from KD with LLaMA-1B and LLaMA-3B RevKD models improving their SFT counterparts by $12.7\%$ and $4.8\%$ respectively.
This reaffirms that although KD effectiveness depends on model size, it does not depend on pre-training differences or reasoning architectures. Overall, these findings underscore that KD is highly effective for commonsense reasoning in smaller models but offers diminishing returns for larger models, necessitating more adaptive distillation strategies for complex reasoning tasks.

On instruction following tasks as well, RevKD outperforms other KD methods. Distilled Qwen-0.5B improves highest on SNI and UNI by 0.11 and 0.15 absolute points respectively. Similarly other distilled Qwen and Llama models achieve largest improvements on these two tasks.

\paragraph{Teacher-student agreement.} Figure~\ref{fig:student_agreement2} and~\ref{fig:student_agreement_llama} highlights the key differences between SFT, KD, and self-agreement across various mathematical and commonsense reasoning tasks for Qwen-14B and LLaMA-8B teacher models, respectively. Generally, KD enhances agreement between the teacher and student compared to SFT, with improvements most pronounced in smaller student models. For instance, in the LLaMA-1B model, agreement in MultiArith improves from $62.5\%$ (SFT) to $71.5\%$ (KD), and in GSM8K from $12.66\%$ to $14.02\%$. Similarly, for Qwen2.5-0.5B distilled from Qwen2.5-14B, KD increases agreement in AddSub ($71.39\%$ to $81.77\%$) and MultiArith ($73.67\%$ to $84.67\%$). However, as student models grow larger, agreement gains diminish, with Qwen2.5-7B showing marginal improvements in GSM8K and SingleEq. Interestingly, self-agreement, which measures alignment between SFT and KD versions of the same student, exhibits lower scores than KD-teacher agreement, suggesting that knowledge distillation introduces distinct learning patterns. For instance, in the Qwen2.5-0.5B model, self-agreement in GSM8K drops from $34.95\%$ (KD) to $20.92\%$, and in MultiArith from $84.67\%$ to $70.67\%$. 

Similarly on commonsense reasoning tasks, KD improves agreement over SFT, particularly in smaller student models. For instance, in the Qwen2.5-0.5B model distilled from Qwen2.5-14B, agreement in ARC-c increases from $50.94\%$ (SFT) to $54.35\%$ (KD) while in ARC-e, it improves from $66.45\%$ to $70.74\%$. However, the impact of KD is less pronounced in larger students, with Qwen2.5-7B showing only marginal gains across most tasks. Interestingly, in higher-complexity datasets like Hellaswag, Winogrande, and BoolQ, KD does not always lead to a significant increase in agreement.

\paragraph{Teacher-student fidelty.} Figure~\ref{fig:student_fidelty2} and~\ref{fig:student_fidelty_llama} illustrates the teacher-student fidelity for Qwen-14B and LLaMA-8B teacher models, respectively. The reasoning fidelity analysis highlights how closely student models replicate their teacher’s reasoning process post-distillation. Across all models, KD improves fidelity over SFT, demonstrating that knowledge transfer enhances reasoning similarity. With Qwen2.5-0.5B model distilled from Qwen2.5-14B, fidelity in GSM8K increases from $47.31\%$ (SFT) to $51.39\%$ (KD), and in SingleEq, it improves from $55.35\%$ to $59.76\%$. Similarly, for the larger Qwen2.5-3B model, fidelity scores increase across tasks, such as MultiArith (from $54.28\%$ to $58.01\%$) and SVAMP (from $51.28\%$ to $55.18\%$). However, the improvements are modest for larger students, suggesting that they develop their own reasoning strategies instead of strictly mimicking the teacher. Similarly as seen in Figure~\ref{fig:student_fidelty_int2} and Figure~\ref{fig:student_fidelty_llama_int}, KD significantly improves fidelity over SFT across all instruction following tasks in all the models. SNI and UNI show the largest improvement among instruction following tasks. 

Self-fidelity, which measures the similarity between a KD-trained student and its SFT-trained counterpart, follows a different trend. While self-fidelity remains high, it is often lower than KD-teacher fidelity, particularly in tasks like GSM8K and AQuA, where the reasoning process shifts post-KD. 
For instance, in Qwen2.5-3B and Qwen2.5-1.5B, self-fidelity is typically $2\%$ lower than KD-teacher fidelity, indicating that the knowledge transfer process alters reasoning dynamics slightly. 

This suggests that while KD effectively aligns student reasoning with the teacher, it also induces modifications in reasoning strategies, particularly in smaller models. Overall, the results indicate that KD enhances reasoning similarity but does not necessarily preserve the teacher’s exact decision-making process.

\section{Error Analysis} \label{appx:error_analysis}
In Table~\ref{tab:error_analysis3}, we show three examples from the validation set of the SVAMP dataset for a more detailed analysis. The student computes an incorrect final answer in the first two examples, while it gets the correct answer in the third example.

Examining the first example, we observe that the student model's intermediate reasoning steps are accurate. The only error is the substitution of ``$37$'' for ``$33$'' in the tower's block count. Despite the incorrect final answer, the student model's reasoning steps are valid and align closely with the teacher model's steps. This alignment is effectively captured by the high `fidelity' score between the student and teacher output.

Similarly, in the second example, the student model computes ``$28 \times 4$'' as ``56'' instead of the correct answer, ``$112$''. However, the intermediate reasoning steps are accurate and align closely with the teacher model's steps, resulting in a high fidelity score. This emphasis on the accuracy of the student model's reasoning steps instills confidence in the model's capabilities.

Conversely, in the third example, we observe that the intermediate reasoning steps of the student and teacher models are entirely different, yet both lead to the correct answer. Here, the final answer score plays a crucial role in evaluating the student's performance. 

\begin{table*}[!htb]
    \captionsetup{font=normalsize}
    \centering
    \scalebox{0.8}{
    \begin{tabular}{cccccccccc}
        \hline
        \textbf{Teacher} & \textbf{Student} & \textbf{Method} & \textbf{GSM8K} & \textbf{SVAMP} & \textbf{MultiArith} & \textbf{SingleEq} & \textbf{AddSub} & \textbf{AQuA} & \textbf{Average} \\ 
        \hline

\multirow{8}{*}{-}
 & Qwen-0.5B & SFT & 23.50 & 41.90 & 76.17 & 77.17 & 74.68 & 23.62 & 52.84 \\
 & Qwen-1.5B & SFT & 49.05 & 66.60 & 89.83 & 89.17 & 84.56 & 25.98 & 67.53 \\
 & Qwen-3B & SFT & 59.29 & 71.80 & 92.67 & 89.57 & 84.81 & 30.71 & 71.47 \\
 & Qwen-7B & SFT & 69.98 & 80.20 & 96.83 & 94.88 & 91.14 & 33.86 & 77.81 \\
 & Qwen-14B & SFT & 75.36 & 82.30 & 96.17 & 94.69 & 89.37 & 38.19 & 79.35 \\
 & Llama-1B & SFT & 12.36 & 29.40 & 64.00 & 63.78 & 63.04 & 19.29 & 41.98 \\
 & Llama-3B & SFT & 39.50 & 53.20 & 90.67 & 86.61 & 86.33 & 23.62 & 63.32 \\
 & Llama-8B & SFT & 58.83 & 69.10 & 94.50 & 91.93 & 85.32 & 29.92 & 71.60 \\
\hline

\multirow{6}{*}{Qwen-3B}
& \multirow{3}{*}{Qwen-0.5B} & SeqKD & 27.45 & 43.90 & 82.5 & 79.13 & 78.23 & 24.02 & 55.87 \\
&  & RevKD & 32.75 & 50.70 & 82.50 & 82.09 & 81.52 & 29.13 & 59.78 \\
&  & GKD & 29.04 & 47.00 & 84.33 & 83.27 & 76.96 & 21.65 & 57.04 \\
 \cdashline{3-10}

& \multirow{3}{*}{Qwen-1.5B} & SeqKD & 42.61 & 63.80 & 86.67 & 87.80 & 82.53 & 27.17 & 65.10 \\
&  & RevKD & 53.37 & 68.80 & 92.17 & 91.14 & 84.81 & 33.46 & 70.62 \\
&  & GKD & 48.14 & 64.70 & 91.67 & 91.34 & 86.08 & 29.92 & 68.64 \\
\hline

\multirow{9}{*}{Qwen-7B}
 & \multirow{3}{*}{Qwen-0.5B} & SeqKD & 27.37 & 46.60 & 83.67 & 81.30 & 76.71 & 25.59 & 56.87 \\
 &  & RevKD & 35.18 & 54.70 & 88.67 & 88.39 & 85.57 & 25.98 & 63.08 \\
 &  & GKD & 33.43 & 52.80 & 90.00 & 83.66 & 82.78 & 24.02 & 61.11 \\
 \cdashline{3-10}

 & \multirow{3}{*}{Qwen-1.5B} & SeqKD & 50.57 & 68.10 & 95.00 & 94.09 & 86.33 & 35.04 & 71.52 \\
 &  & RevKD & 59.21 & 76.70 & 96.67 & 95.87 & 90.38 & 30.71 & 74.92 \\
 &  & GKD & 58.00 & 72.00 & 96.17 & 94.29 & 86.33 & 30.31 & 72.85 \\
  \cdashline{3-10}

 & \multirow{3}{*}{Qwen-3B} & SeqKD & 62.70 & 75.60 & 96.17 & 93.90 & 87.85 & 27.56 & 73.96 \\
 &  & RevKD & 69.90 & 78.70 & 97.33 & 92.52 & 88.35 & 34.65 & 76.91 \\
 &  & GKD & 63.68 & 79.30 & 97.67 & 95.87 & 90.63 & 29.92 & 76.18 \\
\hline

\multirow{12}{*}{Qwen-14B}
 & \multirow{3}{*}{Qwen-0.5B} & SeqKD & 28.43 & 44.6 & 86.17 & 81.50 & 78.23 & 22.44 & 56.90 \\
 &  & RevKD & 35.78 & 53.80 & 87.17 & 87.60 & 83.80 & 31.10 & 63.21 \\
 &  & GKD & 33.66 & 50.50 & 91.17 & 85.04 & 82.28 & 22.44 & 60.85 \\
  \cdashline{3-10}

 & \multirow{3}{*}{Qwen-1.5B} & SeqKD & 50.34 & 66.10 & 93.00 & 91.93 & 87.59 & 30.31 & 69.88 \\
 &  & RevKD & 59.89 & 73.10 & 96.50 & 93.70 & 89.87 & 29.53 & 73.77 \\
 &  & GKD & 55.95 & 72.20 & 95.17 & 92.13 & 86.08 & 31.89 & 72.24 \\
  \cdashline{3-10}

 & \multirow{3}{*}{Qwen-3B} & SeqKD & 61.56 & 73.10 & 96.17 & 93.50 & 87.09 & 29.92 & 73.56 \\
 &  & RevKD & 67.10 & 79.70 & 97.00 & 92.72 & 90.38 & 32.28 & 76.53 \\
 &  & GKD & 63.38 & 77.10 & 96.33 & 95.08 & 89.37 & 34.25 & 75.92 \\
  \cdashline{3-10}

 & \multirow{3}{*}{Qwen-7B} & SeqKD & 70.96 & 78.90 & 97.17 & 95.08 & 90.13 & 30.71 & 77.16 \\
 &  & RevKD & 74.91 & 82.50 & 97.17 & 94.69 & 87.34 & 31.50 & 78.02 \\
 &  & GKD & 72.02 & 80.60 & 96.50 & 94.88 & 88.61 & 34.65 & 77.88 \\

\hline

\multirow{6}{*}{Llama-8B}
 & \multirow{3}{*}{Llama-1B} & SeqKD & 14.33 & 29.50 & 73.17 & 64.76 & 69.87 & 21.26 & 45.48 \\
 &  & RevKD & 20.77 & 36.30 & 77.67 & 69.69 & 71.65 & 24.02 & 50.02 \\
 &  & GKD & 21.68 & 41.70 & 86.33 & 76.77 & 76.46 & 21.26 & 54.03 \\
 \cdashline{3-10}
 & \multirow{3}{*}{Llama-3B} & SeqKD & 38.21 & 55.70 & 91.67 & 87.99 & 85.57 & 22.05 & 63.53 \\
 &  & RevKD & 45.34 & 60.80 & 90.83 & 88.58 & 85.57 & 24.80 & 65.99 \\
 &  & GKD & 45.72 & 61.10 & 94.33 & 90.94 & 85.06 & 24.80 & 66.99 \\

 \hline
    \end{tabular}
    }
    \caption{Performance of different KD methods on mathematical reasoning tasks.}
    \label{tab:math_all_results}
\end{table*}

\begin{table*}[!htb]
    \captionsetup{font=normalsize}
    \centering
    \scalebox{0.7}{
    \begin{tabular}{cccccccccccc}
        \hline
        \textbf{Teacher} & \textbf{Student} & \textbf{Method} & \textbf{Hellaswag} & \textbf{BoolQ} & \textbf{SiQA} & \textbf{PiQA} & \textbf{ARC-e} & \textbf{ARC-c} & \textbf{Winogrande} &  \textbf{OBQA} & \textbf{Average} \\ 
        \hline

\multirow{8}{*}{-}
 & Qwen-0.5B & SFT & 29.90 & 53.91 & 53.84 & 60.50 & 67.00 & 50.94 & 47.91 & 59.40 & 52.92 \\
 & Qwen-1.5B & SFT & 63.90 & 61.28 & 63.25 & 76.93 & 85.40 & 72.18 & 59.51 & 75.00 & 69.68 \\
 & Qwen-3B & SFT & 81.91 & 59.79 & 71.08 & 81.56 & 92.42 & 80.12 & 68.90 & 82.60 & 77.30 \\
 & Qwen-7B & SFT & 91.19 & 66.76 & 74.72 & 87.81 & 94.61 & 87.80 & 74.35 & 89.40 & 83.33 \\
 & Qwen-14B & SFT & 94.39 & 71.47 & 77.99 & 90.70 & 96.93 & 92.41 & 82.79 & 94.60 & 87.66 \\
 & Llama-1B & SFT & 27.37 & 54.40 & 49.74 & 55.60 & 59.60 & 40.87 & 48.22 & 48.40 & 48.02 \\
 & Llama-3B & SFT & 65.01 & 61.99 & 70.37 & 80.47 & 82.91 & 69.03 & 65.98 & 74.20 & 71.25 \\
 & Llama-8B & SFT & 88.89 & 69.33 & 73.03 & 85.26 & 90.15 & 81.66 & 76.87 & 82.80 & 81.00 \\
 \hline

\multirow{6}{*}{Qwen-3B}
 & \multirow{3}{*}{Qwen-0.5B} & SeqKD & 33.66 & 56.42 & 57.68 & 62.30 & 71.97 & 54.44 & 50.99 & 59.60 & 55.88 \\
 &  & RevKD & 39.77 & 56.48 & 59.52 & 67.90 & 74.58 & 56.31 & 55.96 & 66.00 & 59.56 \\
 &  & GKD & 36.45 & 58.75 & 58.96 & 66.87 & 74.41 & 56.40 & 44.20 & 63.60 & 57.45 \\
 \cdashline{3-12}
 & \multirow{3}{*}{Qwen-1.5B} & SeqKD & 66.68 & 59.30 & 56.45 & 78.24 & 82.28 & 71.84 & 63.30 & 72.60 & 68.84 \\
 &  & RevKD & 16.72 & 50.83 & 66.53 & 50.60 & 84.47 & 69.88 & 56.43 & 77.80 & 59.16 \\
 &  & GKD & 72.12 & 58.47 & 69.70 & 79.82 & 89.14 & 76.11 & 63.38 & 81.00 & 73.72 \\
\hline

\multirow{9}{*}{Qwen-7B}
 & \multirow{3}{*}{Qwen-0.5B} & SeqKD & 31.26 & 61.99 & 53.22 & 58.54 & 68.60 & 56.06 & 51.54 & 60.80 & 55.25 \\
 &  & RevKD & 44.13 & 58.04 & 60.29 & 69.15 & 75.17 & 57.34 & 54.54 & 67.20 & 60.73 \\
 
 &  & GKD & 39.42 & 55.72 & 60.80 & 66.87 & 74.49 & 54.61 & 51.54 & 65.40 & 58.61 \\
  \cdashline{3-12}
 & \multirow{3}{*}{Qwen-1.5B} & SeqKD & 70.71 & 60.67 & 66.22 & 73.56 & 87.71 & 73.21 & 54.30 & 81.00 & 70.92 \\
 &  & RevKD & 27.76 & 60.52 & 61.77 & 73.67 & 82.83 & 63.57 & 49.72 & 70.80 & 61.33 \\
 &  & GKD & 76.14 & 60.95 & 70.93 & 81.18 & 89.52 & 76.96 & 65.19 & 82.60 & 75.43 \\
  \cdashline{3-12}
 & \multirow{3}{*}{Qwen-3B} & SeqKD & 81.49 & 63.18 & 71.85 & 82.10 & 93.39 & 83.28 & 69.61 & 86.60 & 78.94 \\
 &  & RevKD & 60.23 & 65.35 & 74.87 & 84.33 & 94.28 & 83.36 & 72.45 & 89.00 & 77.98 \\
 &  & GKD & 11.67 & 57.92 & 74.67 & 81.12 & 93.52 & 82.25 & 71.35 & 85.60 & 69.76 \\
 
\hline 

\multirow{12}{*}{Qwen-14B}
 & \multirow{3}{*}{Qwen-0.5B} & SeqKD & 29.13 & 60.52 & 54.91 & 61.86 & 71.09 & 53.75 & 51.70 & 61.80 & 55.59 \\
 &  & RevKD & 37.27 & 26.79 & 57.88 & 64.04 & 72.52 & 55.55 & 53.75 & 65.20 & 54.12 \\
 &  & GKD & 36.76 & 57.19 & 58.85 & 66.87 & 74.24 & 56.91 & 51.70 & 64.60 & 58.39 \\
  \cdashline{3-12}
 & \multirow{3}{*}{Qwen-1.5B} & SeqKD & 66.70 & 62.20 & 60.70 & 78.02 & 81.31 & 69.88 & 63.14 & 77.80 & 69.97 \\
 &  & RevKD & 72.63 & 63.91 & 71.34 & 82.15 & 89.77 & 77.13 & 63.69 & 82.60 & 75.40 \\
 &  & GKD & 76.59 & 62.42 & 71.14 & 80.63 & 88.80 & 77.22 & 64.33 & 81.80 & 75.37 \\
  \cdashline{3-12}
 & \multirow{3}{*}{Qwen-3B} & SeqKD & 78.64 & 63.15 & 72.88 & 82.64 & 92.89 & 82.25 & 72.14 & 85.60 & 78.77 \\
 &  & RevKD & 86.09 & 65.14 & 75.84 & 84.17 & 93.52 & 83.62 & 72.14 & 85.80 & 80.79 \\
 &  & GKD & 86.57 & 65.23 & 73.85 & 83.08 & 93.77 & 83.28 & 72.69 & 84.60 & 80.38 \\
  \cdashline{3-12}
 & \multirow{3}{*}{Qwen-7B} & SeqKD & 90.18 & 62.87 & 76.15 & 88.19 & 94.53 & 86.52 & 79.16 & 90.80 & 83.55 \\
 &  & RevKD & 92.44 & 69.91 & 79.12 & 88.47 & 95.50 & 88.74 & 78.30 & 89.60 & 85.26 \\
 &  & GKD & 92.60 & 68.93 & 77.38 & 87.81 & 95.20 & 87.88 & 78.77 & 90.60 & 84.90 \\
 \hline

\multirow{6}{*}{Llama-8B}
 & \multirow{3}{*}{Llama-1B} & SeqKD & 29.83 & 54.40 & 57.83 & 62.02 & 65.49 & 47.35 & 52.17 & 57.80 & 53.36 \\
 &  & RevKD & 39.52 & 58.87 & 64.38 & 70.35 & 74.71 & 52.65 & 59.35 & 65.60 & 60.68 \\
 &  & GKD & 39.24 & 57.68 & 62.18 & 69.21 & 73.86 & 51.62 & 57.06 & 63.80 & 59.33 \\
 \cdashline{3-12}
 & \multirow{3}{*}{Llama-3B} & SeqKD & 71.69 & 62.11 & 70.52 & 80.63 & 83.67 & 72.61 & 70.80 & 76.40 & 73.55 \\
 &  & RevKD & 78.08 & 66.70 & 73.08 & 83.30 & 86.11 & 72.61 & 70.56 & 78.20 & 76.08 \\
 &  & GKD & 80.83 & 64.31 & 73.80 & 81.94 & 85.94 & 74.15 & 69.46 & 78.60 & 76.13 \\

        \hline
    \end{tabular}
    }
    \caption{Performance of different KD methods on commonsense reasoning tasks.}
    \label{tab:cs_all_results}
\end{table*}
\begin{table*}[!htb]
    \captionsetup{font=normalsize}
    \centering
    \scalebox{0.8}{
    \begin{tabular}{ccccccccc}
        \hline
        \textbf{Teacher} & \textbf{Student} & \textbf{Method} & \textbf{Dolly} & \textbf{Self} & \textbf{Vicuna} & \textbf{SNI} & \textbf{UNI} & \textbf{Average} \\ 
        \hline

\multirow{8}{*}{-}
 & Qwen-0.5B & SFT & 0.24 & 0.18 & 0.17 & 0.27 & 0.26 & 0.22 \\
 & Qwen-1.5B & SFT & 0.26 & 0.19 & 0.17 & 0.31 & 0.31 & 0.25\\
 & Qwen-3B & SFT & 0.28 & 0.21 & 0.19 & 0.36 & 0.31 & 0.27 \\
 & Qwen-7B & SFT & 0.29 & 0.22 & 0.18 & 0.38 & 0.32 & 0.28 \\
 & Qwen-14B & SFT & 0.30 & 0.25 & 0.19 & 0.41 & 0.36 & 0.30 \\
 & Llama-1B & SFT & 0.24 & 0.14 & 0.16 & 0.26 & 0.28 & 0.22 \\
 & Llama-3B & SFT & 0.27 & 0.19 & 0.17 & 0.31 & 0.33 & 0.25 \\
 & Llama-8B & SFT & 0.30 & 0.22 & 0.19 & 0.31 & 0.35 & 0.27 \\

\hline

\multirow{6}{*}{Qwen-3B}
& \multirow{3}{*}{Qwen-0.5B} 
   & SeqKD & 0.23 & 0.15 & 0.15 & 0.29 & 0.29 & 0.22 \\
&  & RevKD & 0.28 & 0.21 & 0.21 & 0.38 & 0.41 & 0.30 \\
&  & GKD & 0.27 & 0.21 & 0.19 & 0.38 & 0.37 & 0.28 \\
\cdashline{3-9}

& \multirow{3}{*}{Qwen-1.5B} 
  & SeqKD & 0.26 & 0.20 & 0.17 & 0.34 & 0.34 & 0.26 \\
& & RevKD & 0.31 & 0.24 & 0.23 & 0.41 & 0.44 & 0.33\\
& & GKD & 0.29 & 0.24 & 0.20 & 0.39 & 0.39 & 0.30 \\
\hline

\multirow{9}{*}{Qwen-7B}
 & \multirow{3}{*}{Qwen-0.5B} 
     & SeqKD & 0.23 & 0.14 & 0.16 & 0.27 & 0.26 & 0.21 \\
 &   & RevKD & 0.27 & 0.20 & 0.21 & 0.38 & 0.41 & 0.29 \\
 &   & GKD & 0.28 & 0.20 & 0.22 & 0.38 & 0.38 & 0.29 \\
 \cdashline{3-9}

 & \multirow{3}{*}{Qwen-1.5B} 
    & SeqKD & 0.27 & 0.18 & 0.17 & 0.32 & 0.33 & 0.25 \\
&   & RevKD & 0.31 & 0.22 & 0.23 & 0.41 & 0.46 & 0.33 \\
&   & GKD & 0.30 & 0.24 & 0.20 & 0.41 & 0.43 & 0.32 \\
  \cdashline{3-9}

 & \multirow{3}{*}{Qwen-3B} 
    & SeqKD & 0.28 & 0.20 & 0.18 & 0.34 & 0.32 & 0.26 \\
&   & RevKD & 0.33 & 0.25 & 0.23 & 0.42 & 0.46 & 0.34 \\
&   & GKD & 0.32 & 0.25 & 0.22 & 0.40 & 0.42 & 0.32 \\
\hline

\multirow{12}{*}{Qwen-14B}
 & \multirow{3}{*}{Qwen-0.5B} 
     & SeqKD & 0.23 & 0.15 & 0.15 & 0.28 & 0.27 & 0.22 \\
 &   & RevKD & 0.28 & 0.20 & 0.21 & 0.39 & 0.41 & 0.30 \\
 &   & GKD & 0.29 & 0.22 & 0.22 & 0.39 & 0.40 & 0.30 \\
  \cdashline{3-9}

 & \multirow{3}{*}{Qwen-1.5B} 
    & SeqKD & 0.25 & 0.18 & 0.17 & 0.31 & 0.31 & 0.24 \\
 &  & RevKD & 0.32 & 0.25 & 0.23 & 0.43 & 0.45 & 0.34 \\
 &  & GKD & 0.31 & 0.24 & 0.22 & 0.43 & 0.42 & 0.32 \\
  \cdashline{3-9}

 & \multirow{3}{*}{Qwen-3B} 
    & SeqKD & 0.27 & 0.20 & 0.17 & 0.34 & 0.32 & 0.26 \\
 &  & RevKD & 0.32 & 0.23 & 0.23 & 0.43 & 0.46 & 0.33 \\
 &  & GKD & 0.32 & 0.25 & 0.22 & 0.44 & 0.44 & 0.33 \\
  \cdashline{3-9}

 & \multirow{3}{*}{Qwen-7B} 
   & SeqKD & 0.28 & 0.22 & 0.17 & 0.38 & 0.32 & 0.27 \\
 &  & RevKD & 0.33 & 0.28 & 0.23 & 0.48 & 0.47 & 0.36 \\
 &  & GKD & 0.33 & 0.28 & 0.23 & 0.45 & 0.46 & 0.35 \\

\hline

\multirow{6}{*}{Llama-8B}
 & \multirow{3}{*}{Llama-1B} 
    & SeqKD & 0.23 & 0.16 & 0.15 & 0.23 & 0.27 & 0.21 \\
 &  & RevKD & 0.29 & 0.19 & 0.18 & 0.34 & 0.41 & 0.28 \\
 &  & GKD & 0.30 & 0.20 & 0.20 & 0.32 & 0.38 & 0.28 \\
 \cdashline{3-9}
 & \multirow{3}{*}{Llama-3B} 
    & KD & 0.27 & 0.19 & 0.17 & 0.29 & 0.31 & 0.25 \\
 &  & RevKD & 0.33 & 0.24 & 0.21 & 0.39 & 0.44 & 0.32 \\
 &  & GKD & 0.33 & 0.26 & 0.21 & 0.39 & 0.41 & 0.32 \\

 \hline
    \end{tabular}
    }
    \caption{Performance of different KD methods on instruction following tasks using Rouge-L metric.}
    \label{tab:dolly_all_results}
\end{table*}

\begin{table*}[htbp]
  \centering
  \scalebox{0.85}{
    \begin{tabular}
    {|p{10em} | p{13em} | p{13em} | l |}
    \cline{1-4}
    \multicolumn{1}{|l|}{\textbf{Instruction}} & \multicolumn{1}{|l|}{\textbf{Student KD Output}} & \multicolumn{1}{|l|}{\textbf{Teacher SFT Output}} & \multicolumn{1}{|l|}{\textbf{Fidelity}} \\
    \cline{1-4}
    Randy has 37 blocks. He uses 33 blocks to build a tower and 13 blocks to build a house. How many more blocks did he use to build the tower than he did to build the house? & To find the difference between the number of blocks used to build the tower and the house, you can subtract the number of blocks used to build the house from the number of blocks used to build the tower. Here are the steps:                   \colorbox{red}{37 blocks} - 13 blocks = 24 blocks  Therefore, Randy used 24 more blocks to build the tower than he did to build the house.<|endoftext|> & Step 1: Find the total number of blocks used 33 blocks (for the tower) + 13 blocks (for the house) = 46 blocks  Step 2: Find the difference between the number of blocks used for the tower and the house \colorbox{green}{33 blocks} (for the tower) - 13 blocks (for the house) = 20 blocks  Therefore, Randy used 20 more blocks to build the tower than he did to build the house. The answer in Arabic numerals is 20.<|endoftext|> & 0.77 \\
    \cdashline{1-4}
    Robin has 28 packages of gum and 13 packages of candy. There are 4 pieces in each package. How many pieces of gum does Robin have? & To solve this problem, we need to use multiplication.  1. Identify the given information: Robin has 28 packages of gum and 13 packages of candy. Each package has 4 pieces of gum and candy.  2. Determine the operation needed to find the total number of pieces of gum: We need to multiply the number of packages of gum by the number of pieces of gum in each package.  3. Multiply: 28 by 4 to find the total number of pieces of gum.   \colorbox{red}{56. The answer is 56.}  Therefore, Robin has 56 pieces of gum.<|endoftext|> & To solve this problem, we need to use multiplication. 1. Multiply the number of packages of gum by the number of pieces in each package:  \colorbox{green}{28 packages x 4 pieces = 112} pieces of gum  2. Therefore, Robin has 112 pieces of gum.  The answer in Arabic numerals is: 112.<|endoftext|> & 0.79 \\
    \cdashline{1-4}
Because of the decision Sofia asked 310 students to suggest specific types of food. 185 students suggested adding mashed potatoes while others suggested adding bacon to the menu. How many students suggested bacon? & To solve the problem, we can use subtraction:Total number of students = 310Number of students who suggested mashed potatoes = 185To find the number of students who suggested bacon, we subtract the number of students who suggested mashed potatoes from the total number of students:  \colorbox{green}{310 - 185 = 125}  Therefore, 125 students suggested adding bacon to the menu.<|endoftext|> & We can start by using the concept of complementary counting. Let x be the number of students who suggested bacon. We know that the total number of students is 310, so we can set up an equation: 185 + x = 310 Subtracting 185 from both sides: \colorbox{green}{x = 310 - 185} Simplifying:  x = 125 Therefore, 125 students suggested adding bacon to the menu. Answer: 125.<|endoftext|> & 0.13 \\
    \cline{1-4}
    \end{tabular}%
    }
    \caption{Error analysis of Qwen student models on SVAMP validation data.}
    \label{tab:error_analysis3}%
\end{table*}%


\end{document}